\documentclass[journal]{IEEEtran}
\usepackage{graphicx}
\usepackage{multirow}
\usepackage{epstopdf}
\usepackage{times}
\usepackage{amsmath}
\usepackage{amssymb}
\usepackage{amsfonts}
\usepackage{color}
\usepackage{subfigure}
\usepackage{hyperref}

\usepackage{algorithm}
\usepackage{algorithmic}
\usepackage{rotating}

\usepackage{adjustbox,lipsum}
\usepackage{dsfont}
\usepackage{booktabs}
\usepackage{multirow}
\usepackage{array}

\begin{document}

\title{Progressive Cross-camera Soft-label Learning for Semi-supervised  Person Re-identification}

\author{Lei Qi,
        Lei Wang,
        Jing Huo,
        Yinghuan Shi,
        Yang Gao
\thanks{Lei Qi, Jing Huo, Yinghuan Shi and Yang Gao (corresponding author) are with the State Key Laboratory for Novel Software Technology, Nanjing University, Nanjing, China, 210023 (e-mail: qilei.cs@gmail.com; huojing@nju.edu.cn; syh@nju.edu.cn; gaoy@nju.edu.cn).}
\thanks{Lei Wang is School of Computing and Information Technology, University of Wollongong, Wollongong, Australia (e-mail: leiw@uow.edu.au).}

}

%
%

\markboth{IEEE TRANSACTIONS ON CIRCUITS AND SYSTEMS FOR VIDEO TECHNOLOGY,}%
{Shell \MakeLowercase{\textit{et al.}}: Bare Demo of IEEEtran.cls for IEEE Journals}

\maketitle

\begin{abstract}
 In this paper, we focus on the semi-supervised person re-identification (Re-ID) case, which only has the intra-camera (within-camera) labels but not inter-camera (cross-camera) labels. In real-world applications, these intra-camera labels can be readily captured by tracking algorithms or few manual annotations, when compared with cross-camera labels. In this case, it is very difficult to explore the relationships between cross-camera persons in the training stage due to the lack of cross-camera label information.
 To deal with this issue, we propose a novel Progressive Cross-camera Soft-label Learning (PCSL) framework for the semi-supervised person Re-ID task, which can generate cross-camera soft-labels and utilize them to optimize the network. 
 Concretely, we calculate an affinity matrix based on person-level features and adapt them to produce the similarities between cross-camera persons (i.e., cross-camera soft-labels). 
 To exploit these soft-labels to train the network, we investigate the weighted cross-entropy loss and the weighted triplet loss from the classification and discrimination perspectives, respectively. Particularly, the proposed framework alternately generates progressive cross-camera soft-labels and gradually improves feature representations in the whole learning course.
Extensive experiments on five large-scale benchmark datasets show that PCSL significantly outperforms the state-of-the-art unsupervised methods that employ labeled source domains or the images generated by the GANs-based models. Furthermore, the proposed method even has a competitive performance with respect to deep supervised Re-ID methods.
\end{abstract}


%
\IEEEpeerreviewmaketitle

\section{Introduction}
\IEEEPARstart{I}{n} recent years, person re-identification (Re-ID) has drawn an increasing interest in both academia and industry due to its great potentials in video analysis and understanding. The person Re-ID aims at matching images of the same person captured by different cameras with non-overlapping camera views~\cite{DBLP:journals/tmm/ChenLLCH11}. This main challenge of person Re-ID is the variations such as body pose, viewing angle, illumination, image resolution, occlusion and background across different cameras. Generally, person Re-ID can be treated as a special case of the image retrieval problem with the goal of querying from a large-scale gallery set to quickly and accurately find images that match with a query image~\cite{DBLP:journals/tip/ZhangLZZZ15}\cite{DBLP:journals/tmm/Wang0TL13}.

\begin{figure}
\centering
\subfigure[Images]{
\includegraphics[width=4cm]{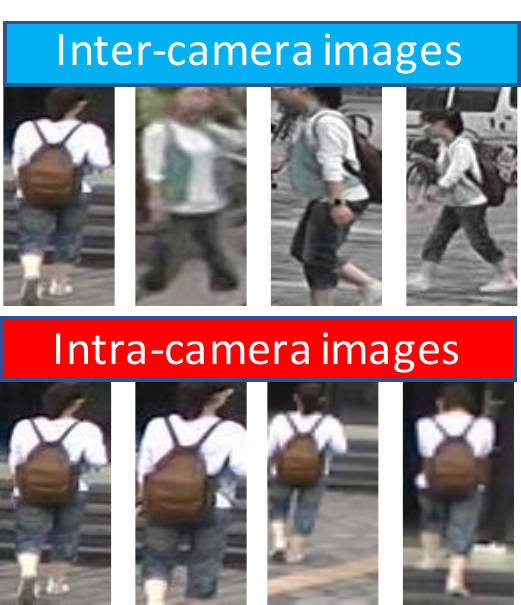}
}
\subfigure[Performance]{
\includegraphics[width=4cm]{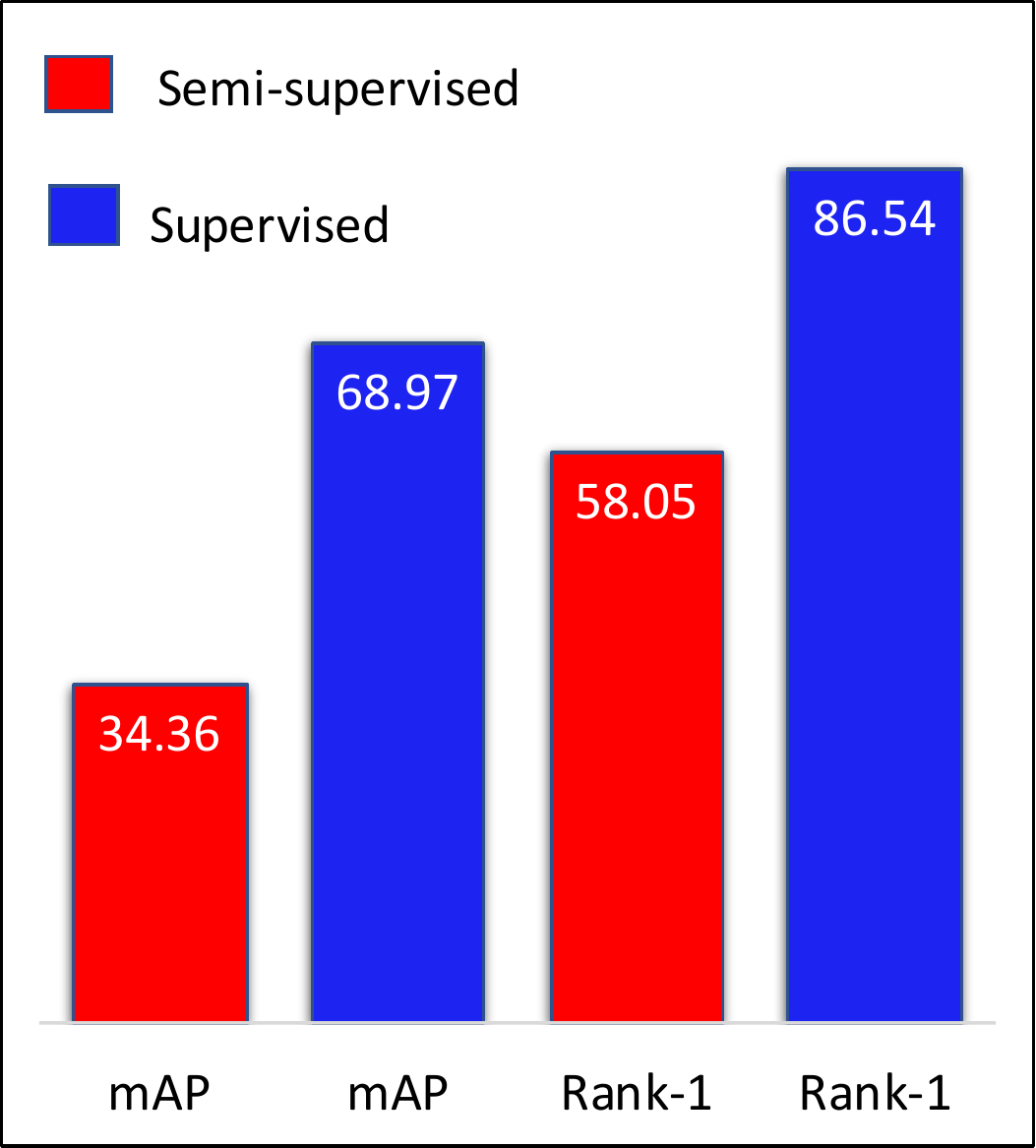}
}
\caption{Comparison between inter-camera and intra-camera images on Market1501~\cite{DBLP:conf/iccv/ZhengSTWWT15}. In detail, a) shows the large discrepancy between inter-camera images and minor variation between intra-camera images; b) is the performance in the semi-supervised and supervised settings, respectively. As seen, if we merely use intra-camera labels to train models, the performance still has a large margin with the supervised method.}
\label{fig5}
\vspace{-15pt}%
\end{figure}

Currently, most existing work for person Re-ID mainly focuses on the supervised~\cite{DBLP:conf/cvpr/XiaoLOW16,DBLP:conf/aaai/ChenCZH17,sun2018beyond,zheng2018re} and unsupervised~\cite{fan2018unsupervised,zhong2018generalizing,Bak_2018_ECCV,wang2018transferable,lv2018unsupervised} cases. Although the supervised person Re-ID methods can achieve good performance in many public datasets, they need large-scale labeled data to train models, especially for the deep-learning-based methods. However, labeling data is very expensive in the person Re-ID community, especially for finding the same identity across camera views. Therefore, supervised methods have poor scalability in real-world scenarios. In the unsupervised person Re-ID task, due to the lack of labeled data for training models, most methods still have a big performance margin when compared with the supervised counterparts~\cite{han2018attribute,Zhao_2017_ICCV,su2017pose,zheng2019pose,zhao2017spindle}. 

In this paper, different from supervised or unsupervised cases, we focus on a ``semi-supervised'' person Re-ID task, in which there are the within-camera (intra-camera) labels but no cross-camera (inter-camera)  label information which are expensive to obtain. In practice, it is easy to obtain the within-camera labels by tracking algorithms~\cite{dehghan2015gmmcp,maksai2017non} and a few manual labeling. Therefore, it is very meaningful to deal with the semi-supervised person Re-ID case in real-world applications. In this case, we aim to utilize these intra-camera labels and explore the cross-camera underlying discriminative information to reduce the performance gap when compared with deep supervised person Re-ID methods and bring a significant improvement over the unsupervised methods that utilize extra labeled source domains or the images obtained from the GANs-based models.

In real-world scenarios, a person's appearance often varies greatly across camera views due to the changes in body pose, view angle, occlusion, image resolution, illumination conditions, and background noises, as shown in Fig.~\ref{fig5} (a). Compared with the intra-camera images, the inter-camera images have even more variations. However, if we only utilize the intra-camera labels in the training stage, the performance still has a large margin with the supervised method (e.g., A$^3$M~\cite{han2018attribute}), as reported in Fig.~\ref{fig5} (b). Therefore, this is a key challenge (i.e., the lack of the inter-camera labels) to handle in the semi-supervised person Re-ID task. 

Currently, several methods are developed to deal with the lack of label information. In the unsupervised person Re-ID task, to solve the problem, assigning pseudo-labels for unlabeled data is introduced in the literature~\cite{fan2018unsupervised,zhong2018generalizing,Bak_2018_ECCV,wang2018transferable,lv2018unsupervised}. Moreover, in~\cite{DBLP:conf/bmvc/LinLLK18,DBLP:journals/corr/abs-1904-01308,qi2019novel}, other labeled datasets (e.g., source domains) or additional images (e.g., images generated by the GAN-based models) are utilized to enhance the generalization ability of models in the unlabeled target domain. However, it is worth noting that in our semi-supervised Re-ID case, we do not employ other labeled datasets and extra images from the GAN-based models to help the semi-supervised learning. Besides, deep semi-supervised learning mainly aims at designing unsupervised loss functions~\cite{sajjadi2016regularization,rasmus2015semi,tarvainen2017mean,qiao2018deep} and generating high-quality pseudo-labels for unlabeled samples~\cite{lee2013pseudo,shi2018transductive,iscen2019label}. However, in our task, we aim to explore the relationships between cross-camera persons to introduce cross-camera discriminative information.

In this paper, to sufficiently mine the discriminative information in unlabeled cross-camera data, we develop a Progressive Cross-camera Soft-label Learning (PCSL) framework for the
semi-supervised person re-identification task, which involves the supervised intra-camera learning and the unsupervised inter-camera learning. For the supervised intra-camera learning, we directly utilize the existing triplet loss function~\cite{hermans2017defense} to train the feature extractor. For the unsupervised inter-camera learning, we exploit an affinity matrix to construct the similarities between cross-camera persons, which are called ``cross-camera soft-labels'' in this paper. 
 To utilize this inter-camera soft-label information, we employ two different schemes to train the proposed network from the classification and discrimination views, which include the weighted cross-entropy loss and the weighted triplet loss. Particularly, generating soft-labels and learning feature representations are iterative in the whole training course and thus it can obtain progressive cross-camera soft-labels and increasingly improved feature representations. We conduct experiments to validate the effectiveness of the proposed method on five large-scale datasets including three image datasets and two video datasets, which shows that the proposed PCSL has a significant improvement when compared with the state-of-the-art unsupervised Re-ID methods that could utilize some extended information. Moreover, PCSL also outperforms many deep supervised person Re-ID methods.

In summary, we propose a novel end-to-end deep network to address the semi-supervised person Re-ID case. The main contributions of this paper are threefold. 
Firstly, we put forward a progressive cross-camera soft-label learning framework for semi-supervised person Re-ID, which alternatively updates cross-camera soft-labels and learns feature representations to smoothly promote both of them in the training process.
Secondly, based on the classification and discrimination perspectives, we respectively exploit the weighted cross-entropy loss and the weighted triplet loss to train the network, which can take full advantage of generated cross-camera soft-labels to mine the discriminative information from the unlabeled inter-camera data. 
Lastly, extensive experiments show the proposed network (i.e., PCSL) significantly outperforms the unsupervised methods that take advantage of the extended data. Moreover, PCSL has comparable results with respect to many deep supervised methods. We also confirm the efficacy of all components in our framework by ablation studies.

The rest of this paper is organized as follows.
We review some related work in Section \ref{s-related}.
The PCSL framework is introduced in Section \ref{s-framework}.
Experimental results and analysis are presented in Section \ref{s-experiment},
and Section \ref{s-conclusion} is conclusion.

\section{Related work}\label{s-related}

\subsection{Unsupervised Person Re-ID}
In the early years, several methods for person Re-ID mainly aim at designing discriminative hand-crafted features~\cite{liao2015person,DBLP:conf/iccv/ZhengSTWWT15}, which can be directly utilized to measure the similarity between two samples. Besides, some domain adaptation methods~\cite{DBLP:conf/cvpr/PengXWPGHT16, DBLP:journals/tcsv/WangZLZ16, DBLP:journals/tip/MaLYL15,qi2018unsupervised,DBLP:conf/iccv/YuWZ17} based on hand-crafted features are developed to transfer the discriminative information from the labeled source domain to the unlabeled target domain by learning a shared subspace or dictionary between source and target domains. However, these methods are not based on deep frameworks and thus do not fully explore the high-level semantics in images.

To overcome the aforementioned issue, many deep unsupervised Re-ID methods are proposed in recent years. Due to the lack of the label information, some methods assign pseudo-labels for unlabeled samples to train deep networks~\cite{fan2018unsupervised,zhong2018generalizing,Bak_2018_ECCV,wang2018transferable,lv2018unsupervised}. In~\cite{fan2018unsupervised}, the clustering method is utilized to give pseudo-labels to the unlabeled samples. Lv \textit{et al.}~\cite{lv2018unsupervised} use the spatio-temporal patterns of pedestrians to obtain more stable pseudo-labels. 
Besides, deep domain adaptation methods are developed to reduce the discrepancy between source and target domains from the feature representation view~\cite{DBLP:conf/bmvc/LinLLK18,DBLP:journals/corr/abs-1904-01308,qi2019novel}.
Delorme \emph{et al.}~\cite{DBLP:journals/corr/abs-1904-01308} introduce an adversarial framework in which the discrepancy across cameras is relieved by fooling a camera discriminator.
Qi \emph{et al.}~\cite{qi2019novel} develop a camera-aware domain adaptation to reduce the discrepancy not only between source and target domains but also across cameras.

Recently, 
enriching training images by the GAN-based models gains a lot of attention~\cite{wei2018person,deng2018image,zhong2018generalizing,Bak_2018_ECCV} in unsupervised Re-ID. 
Wei \textit{et al.}~\cite{wei2018person} impose constraints to maintain the identity in image generation. The approach in~\cite{deng2018image} enforces the self-similarity of an image before and after translation and the domain-dissimilarity of a translated source image and a target image.
Tian \emph{et al.}~\cite{tian2019imitating} exploit the underlying commonality across different domains from the class-style space to improve the generalization ability of Re-ID models.
Zhong \emph{et al.}~\cite{zhong2019invariance} introduce the camera-invariance into the model, which requires that each real image and its style-transferred counterparts share the same identity.

Particularly, the aforementioned methods are based on image data. The unsupervised learning has also attracted increasing interest in video-based person Re-ID~\cite{kodirov2016person,khan2016unsupervised,ye2018robust,liu2017stepwise,ye2017dynamic}. 
Chen \emph{et al.}~\cite{DBLP:conf/bmvc/ChenZG18} learn a deep Re-ID matching model by jointly optimizing two margin-based association losses in an end-to-end manner, which effectively constrains the association of each frame to the best-matched intra-camera representation and cross-camera representation.
Li \emph{et al.}~\cite{Li_2018_ECCV} jointly learn within-camera tracklet association and cross-camera tracklet correlation by maximizing the discovery of most likely tracklet relationships across camera views. Moreover, its extension is capable of incrementally discovering and exploiting the underlying Re-ID discriminative information from automatically generated person tracklet data~\cite{li2019unsupervised}. 

Different from the above pseudo-label methods, which generate pseudo-label for each image (i.e., the image-level label), we aim to produce soft-labels to construct the connection for cross-camera persons (i.e., the person-level label). Besides, our proposed method does not utilize additional data, such as the images generated by GANs or labeled source domains. 

\subsection{Supervised Person Re-ID}
In supervised person Re-ID, most existing methods aim to design the robust features~\cite{DBLP:conf/iccv/ZhengSTWWT15,DBLP:conf/cvpr/MatsukawaOSS16} and learning the discriminative subspace~\cite{DBLP:conf/cvpr/ZhangXG16,koestinger2012large,liao2015person,chen2016asymmetric} before the popularization of deep learning. Recently, deep learning has obtained a huge achievement in person Re-ID~\cite{DBLP:conf/cvpr/XiaoLOW16,sun2017svdnet}. In~\cite{wu2016personnet,DBLP:conf/eccv/VariorHW16,DBLP:conf/eccv/VariorSLXW16}, the networks take an image pair as input, and output a similarity indicating whether the two input images depict the same person. 
Besides, several effective loss functions~\cite{Ristani_2018_CVPR,wu2018deep} are developed to promote the performance.
Ristani \emph{et al.}~\cite{Ristani_2018_CVPR} introduce an adaptive weighted triplet loss for training and a new technique for hard-identity mining.
Moreover, with the help of human information (e.g., attribute or pose), deep Re-ID frameworks can relieve human misalignments and pose variations~\cite{han2018attribute,Zhao_2017_ICCV,su2017pose,zheng2019pose,zhao2017spindle}. 
Recently, the local-based methods~\cite{li2017learning,liu2017stepwise} further improve person Re-ID performance. 
The method in~\cite{liu2017stepwise} jointly learns local and global features in a CNN by discovering correlated local and global feature representations in a different context.

However, these aforementioned supervised person Re-ID methods need a large number of labeled data, which is expensive and unrealistic. Therefore, they have weak scalability in real-world applications.

\subsection{Deep Semi-supervised Learning}
Most existing deep semi-supervised learning (SSL) can be divided into two main categories. First category is designing unsupervised loss in deep SSL~\cite{sajjadi2016regularization,rasmus2015semi,tarvainen2017mean,qiao2018deep}. For example, Sajjadi \emph{et al.}~\cite{sajjadi2016regularization} assume that every training image, labeled or not, belongs to a single category, a natural requirement on the classifier is to make a confident prediction on the training set by minimizing the entropy of the network output.
Besides, \cite{sajjadi2016mutual,DBLP:conf/iclr/LaineA17,qiao2018deep,miyato2018virtual} focus on consistency loss, where two related cases, e.g., coming from two similar images, or made by two networks with related parameters, are encouraged to have similar network outputs. 

The second category is assigning pseudo-labels to the unlabeled examples~\cite{lee2013pseudo,shi2018transductive,iscen2019label}. The method in~\cite{lee2013pseudo} uses the current network to infer pseudo-labels of unlabeled examples, by choosing the most confident class. These pseudo-labels are treated like human-provided labels in the cross-entropy loss. The same principle is adopted by Shi \emph{et al.}~\cite{shi2018transductive}, where the authors further add contrastive loss to the consistency loss. 
Differently, Iscen \emph{et al.}~\cite{iscen2019label} employ a transductive label propagation method that is based on the manifold assumption to make predictions on the entire dataset and use these predictions to generate pseudo-labels for the unlabeled data and train a deep neural network. 

In contrast to deep SSL, which constructs the relationship between different samples, our proposed method for semi-supervised person Re-ID aims to explore the relationship between inter-camera persons.

\section{The proposed method}\label{s-framework}
\begin{figure*}
\centering
\includegraphics[width=18cm]{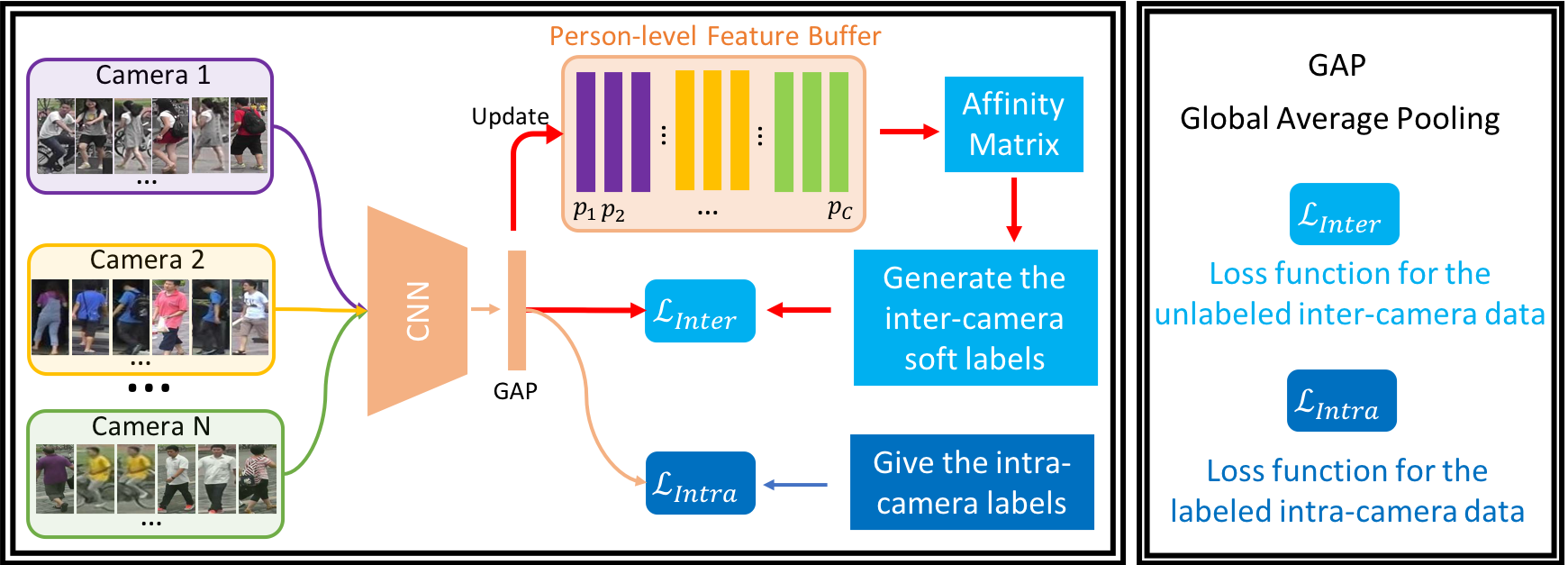}
\caption{An illustration of the proposed progressive cross-camera soft-label learning framework. It consists of two tasks including the supervised intra-camera task and unsupervised inter-camera task. For supervised intra-camera learning, we directly use the given intra-camera labels to train the network. For the unsupervised inter-camera learning, we employ a person-level feature buffer to store the person-level feature representations, which is updated at each iteration. By an affinity matrix computed according to the person-level feature buffer, we can obtain the similarities between the cross-camera persons (i.e., cross-camera soft-labels). Then we exploit these cross-camera soft-labels to train the network. Particularly, progressive cross-camera soft-labels and feature representations can be generated by the alternative way in the whole training stage. Best viewed in color.}
\label{fig1}
\vspace*{-15pt}%
\end{figure*}

In this paper, different from the typical supervised and unsupervised person Re-ID cases, we put forward a novel Progressive Cross-camera Soft-label Learning (PCSL) framework for semi-supervised person Re-ID where only the intra-camera (within-camera) labels are given but no inter-camera (cross-camera) labels are available. We illustrate the proposed method in Fig.~\ref{fig1}, which consists of the supervised intra-camera learning and the unsupervised inter-camera learning. Firstly, for learning the discriminative information from the intra-camera labeled data, we employ the triplet loss with the hard-sample-mining scheme to train the feature extractor~\cite{hermans2017defense}.  
Secondly, for unlabeled inter-camera learning, we develop a cross-camera soft-label learning method, which can generate cross-camera soft-labels to train the proposed network. 
This allows us to fully take advantage of the steadily improved feature representation to produce better soft-labels.
In turn, these soft-labels help networks to learn better feature representations, forming a positive loop to exploit discriminative information from unsupervised inter-camera data.
In particular, we investigate two different loss functions to learn the feature representation with these soft-labels from the classification and discrimination perspectives, respectively.
In the following, the problem definition is firstly presented in Section~\ref{SEC:PD}. Then, the supervised intra-camera learning and the unsupervised inter-camera learning are elaborated in Section~\ref{SEC:SIL} and Section~\ref{SEC:UIL}, respectively. Finally, the optimization of PCSL is described in Section~\ref{SEC:OPT}.

\subsection{Problem Definition}~\label{SEC:PD}
We assume a collection of $n$ cameras $\mathrm{X} = \left \{ \mathrm{X}_1, \mathrm{X}_2, \cdots , \mathrm{X}_n  \right \}$ and $\mathrm{Y}_{Intra} = \left \{\mathrm{Y}_1, \mathrm{Y}_2, \cdots , \mathrm{Y}_n  \right \}$ is its corresponding intra-camera label set, where $\mathrm{X}_i$ and $\mathrm{Y}_i$ are the sample set and the label set in the $i$-th camera. $x_i\in \chi$ is the $i$-th sample in $\mathrm{X}$.  We presume that there are $C_i$ persons (i.e., identities (IDs)) in the $i$-th camera, and $C=C_1+\cdots+C_n$ is the number of IDs in total. Note that, IDs in different cameras may be overlapping, i.e., the same identity could appear in multiple cameras. The goal in semi-supervised person Re-ID is to use all examples $\mathrm{X}$ and the intra-camera labels $\mathrm{Y}_{Intra}$  to train a robust feature representation network $\phi _{\theta}:\chi\rightarrow \mathbb{R}^d$ mapping the input to a feature vector, or descriptor, where $\theta$ is the network parameter and $d$ denotes the feature dimension. We define the descriptor of the $i$-th example $x_i$ by $v_i=\phi_{\theta}(x_i)$.

\subsection{Supervised Intra-camera Learning}\label{SEC:SIL}
In the proposed framework, we first use the sample set $\mathrm{X}$ and the intra-camera label set $\mathrm{Y}_{Intra}$ to train the network. Particularly, we employ the ResNet-50~\cite{DBLP:conf/cvpr/HeZRS16} pre-trained on ImageNet~\cite{DBLP:conf/cvpr/DengDSLL009} as the backbone network and triplet loss as loss function. In this task, since there is no label information across cameras, we only choose training data from the same camera to generate triplets at each iteration. 
Consequently, in each training batch, we randomly select $N_P$ persons and each person has $N_K$ images, i.e., the total number of images is $N=N_P\times N_K$ in a batch. To produce high-quality triplets, we utilize the hard-sample-mining scheme whose effectiveness has been sufficiently validated in the literature~\cite{hermans2017defense}. For the $i$-th camera, the triplet loss with hard sample mining can be described as
\begin{equation}\label{eq01}
\begin{aligned}
\mathcal{L}_{\mathrm{Intra}}=\underbrace{\sum_{i=1}^{N_P}\sum_{a=1}^{N_K}}_{one~batch}[m+l(v_{a}^{i})]_{+},
  \end{aligned}
\end{equation} where $[\cdot]_{+}$ equals to $\max(\cdot, 0)$. For an anchor sample feature $v_{a}^{i}$ belonging to the $i$-th person, 
\begin{equation}\label{eq02}
\begin{aligned}
l(v_{a}^{i})=\overbrace{\max_{p=1...N_K}D(v_{a}^{i}, v_{p}^{i})}^{hardest~positive}-\overbrace{\min_{\substack{j=1...N_P\\ n=1,...N_K \\j\neq i}}D(v_{a}^{i}, v_{n}^{j})}^{hardest~negative},
  \end{aligned}
\end{equation} and $m$ denotes the margin. $D(\cdot,\cdot)$ indicates the Euclidean distance of two feature vectors. In Eq.~(\ref{eq01}), the first term  (second term) is the Euclidean distance of the hardest positive (negative) sample and the anchor.

As described and demonstrated by experiments in~\cite{hermans2017defense}, the triplet loss with the hard-sample mining scheme well optimizes the embedding space such that data points with the same identity are closer to each other than those with different identities in the within-camera view. And this embedding space will be further improved as follows.

\subsection{Unsupervised Inter-camera Learning} \label{SEC:UIL}
In the semi-supervised person Re-ID case, the main challenge is the unsupervised inter-camera learning due to no cross-camera labels are available. In this paper, we develop a novel end-to-end learning framework that can obtain progressive cross-camera soft-labels and gradually improve feature representation learning in the whole training course. The following part is a detailed description of this unsupervised inter-camera learning.

\textbf{Person-level feature buffer.}
 In the proposed framework, we first compute the feature representation for each person. Different from image-level feature representation (i.e., one feature vector represents an image), we need to use a feature vector to represent multiple images involving the same person in the same camera. In this paper, we follow the technique developed to represent tracklets in video-based person Re-ID~\cite{li2019unsupervised,DBLP:conf/bmvc/ChenZG18}, which accumulates the frame features of the same tracklet to represent the tracklet. In this paper, we employ the similar scheme to produce 
a person-level feature and store them in a buffer $\mathbf{P}=\left[p_1, p_2, ..., p_C\right]$, where $p_i$ is the feature of the $i$-th person. At the $t$-th iteration, the feature representation of the $i$-th person is updated as 
 \begin{equation}\label{eq03}
\begin{aligned}
p_i^{t+1}=\frac{1}{2}(p_i^t+\frac{1}{N_K}\sum_{j=1}^{N_K} v_j^i),~~ i \in \left \{1, \cdots, C  \right \}.
  \end{aligned}
\end{equation} 

\emph{Remarks.} The person-level feature buffer can well represent persons due to two reasons. i) Updating the feature buffer at each iteration can guarantee that the buffer always stores the steadily improved feature representations in the whole training process. ii) Our proposed scheme (i.e., Eq.~(\ref{eq03})) is similar to the multi-query scheme (i.e., a query feature generated by merging multiple image features with the same IDs) in person Re-ID. For example, on the Market1501 dataset, which includes two evaluation protocols (i.e., single-query and multi-query), the experiments have fully validated the effectiveness of the multi-query scheme in~\cite{li2017learning,qi2018maskreid}, i.e., the multi-image feature is more robust than the single-image feature. 

\begin{figure}
\centering
\includegraphics[width=8cm]{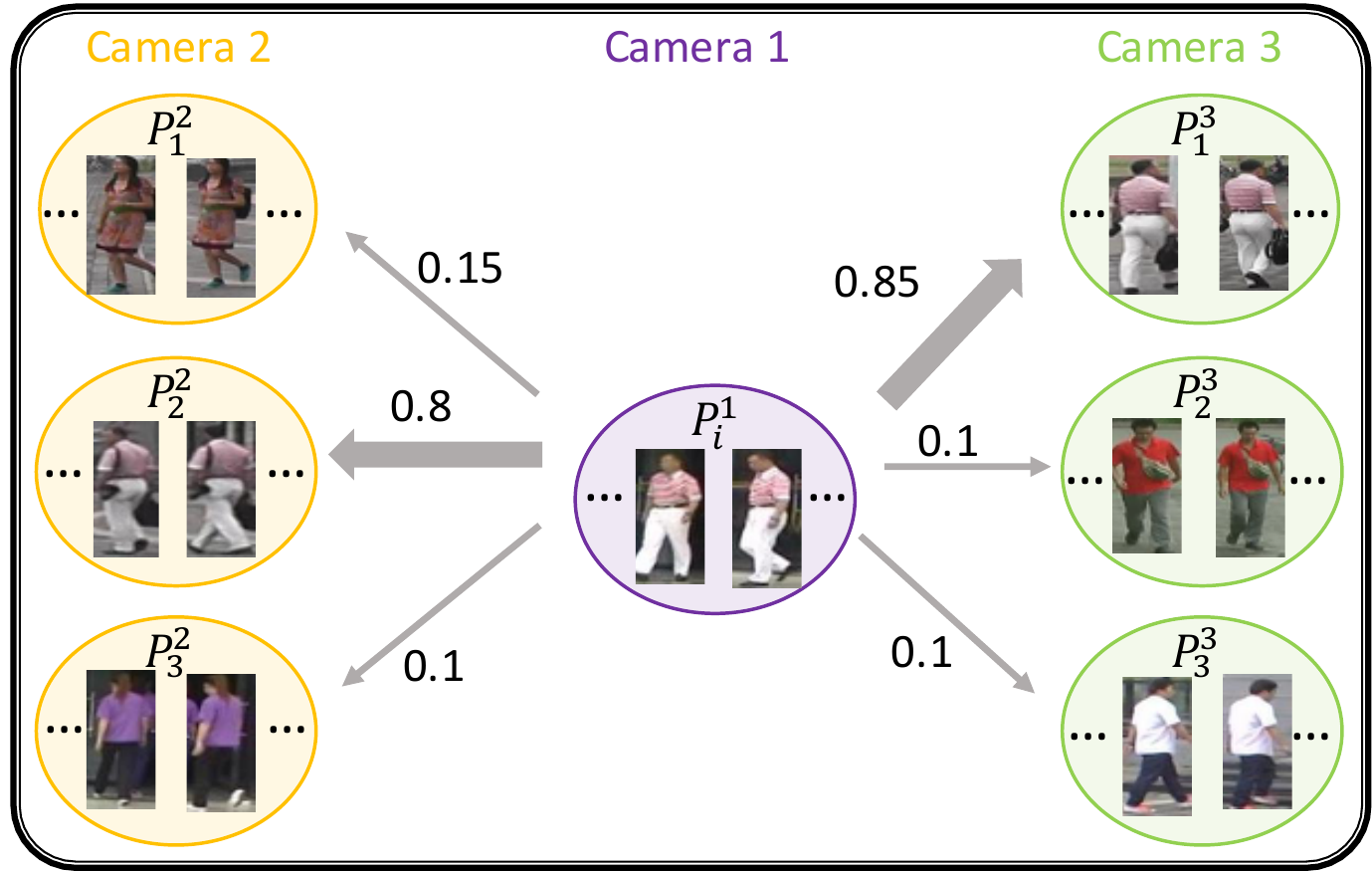}
\caption{An illustration of the cross-camera soft-label. In this figure, $P_i^j$ denotes the $i$-th person in the $j$-th camera. In each camera, since the semi-supervised case has the intra-camera label information, we know that which images belong to $P_i^j$. For clarity, we only show the similarities of the $i$-th person in the $1$-st camera and six persons in two other cameras.
Note that the similarities between cross-camera persons are from the affinity matrix $\mathbf{A}$, which are called ``cross-camera soft-labels'' in this paper. If the similarity of two cross-camera persons is large, this means that they have a large chance that they belong to the true same identity. Best viewed in color.}
\label{fig9}
\vspace*{-15pt}%
\end{figure}

  \textbf{Cross-camera soft-labels.} Then to create the correlation of IDs across different cameras, we propose to generate soft-labels based on the  aforementioned person-level features. Specifically, we compute the affinity matrix $\mathbf{A}\in \mathbb{R}^{C \times C}$ according to the person-level feature buffer $\mathbf{P}\in\mathbb{R}^{d \times C}$, and $a_{i,j}$ denotes the element in $i$-th row and $j$-th column of $\mathbf{A}$, which means the similarity between the $i$-th and $j$-th persons. For the construction of the affinity matrix, we consider that it should have the following two properties: i) Neglecting the other persons in the same camera (i.e., only considering the relationships between cross-camera persons) because the intra-camera labels (i.e., person IDs) are non-overlapping. Otherwise, some extra noises could be brought into the affinity matrix.
ii) Retaining the similarities of the $k$ nearest persons, because for one identity, most other identities are not related with it. Therefore, by employing Gaussian Kernel, we can get affinity matrix as
 \begin{equation}\label{eq04}
\begin{aligned}
a_{i,j}=\left\{\begin{matrix}
\exp (-\frac{{\left \| p_i-p_j \right \|}_{2}^{2}}{\sigma ^{2}}), & if~ \phi_c(p_i) \neq \phi_c(p_j) \wedge  p_i \in \mathcal{N}_k(p_j) \\ 
 0,& otherwise
\end{matrix}\right.
\end{aligned}
\end{equation} where $\phi_{c}(p_i)$ denotes the camera ID of $p_i$. $\mathcal{N}_k(p_j)$ indicates the set of $k$ nearest neighbors of $p_j$. Particularly, as reported in the literature~\cite{li2019unsupervised}, to incorporate the local density structure, we set $\sigma ^{2}$ as the mean value of all elements in the affinity matrix $\mathbf{A}$. The $i$-th row in affinity matrix represents the similarities between the $i$-th person and others, which are called ``cross-camera soft-labels'' in this paper, as shown in Fig.~\ref{fig9}.

In person Re-ID, the cross-entropy loss~\cite{sun2018beyond,Zhao_2017_ICCV} or the triplet loss~\cite{hermans2017defense,Ristani_2018_CVPR} is usually used to train models from the classification or discrimination perspective. Based on cross-camera soft-labels, we investigate the weighted cross-entropy loss and the weighted triplet loss to train our proposed network with the inter-camera soft-label information for the two perspectives, respectively.
\subsubsection{Classification Task}
In this paper, we conduct the classification task to train the network with generated cross-camera soft-labels. Particularly, there are $C$ classes (i.e., the total number of persons in all cameras) in this task. Particularly, these classes are overlapping with each other. i.e., different IDs across different cameras could have the true same identity. The concrete description is given in the following.

 \textbf{Classifier.} The classifier typically consists of a fully connected (FC) layer applied on top of $\phi_{\theta}$, producing a vector of confidence scores. Function $f_{\theta}$ is the mapping from input space directly to confidence scores $y=[y_1, \cdots, y_C]=f_{\theta}(v_i)$. Then we use softmax to normalize the output of the network and produce the confidence score for a sample $x_i$ belonging to the $c$-th class as
\begin{equation}\label{eq05}
\begin{aligned}
P(c|x_i)= \frac{\exp (y_c)}{\sum_{j=1}^{C}\exp (y_j)}.
  \end{aligned}
\end{equation} 

 \textbf{Weighted cross-entropy loss.} 
 The traditional cross-entropy loss only considers one class in the optimization as
\begin{equation}\label{eq11}
\begin{aligned}
&\mathcal{L}_\mathrm{C}= -\sum_{c=1}^{C}\delta(z-c)\log P(c|x_i),
  \end{aligned}
\end{equation} where $\delta(\cdot)$ is the Dirac delta function and $z$ means that the sample $x_i$ is from the $z$-th person. 
Differently, the semi-supervised person Re-ID case exists repetitive classes across cameras, i.e., one sample could belong to multiple classes. For each class, we give a weighted value as 
\begin{equation}\label{eq06}
\begin{aligned}
W_i=\left [\frac{a_{i,1}}{\sum_{j=1}^{C} a_{i,j}}, \cdots, \frac{a_{i,C}}{\sum_{j=1}^{C} a_{i,j}} \right ],~~ i =1, \cdots, C,
  \end{aligned}
\end{equation} where each element represents the probability that one sample belongs to other classes with respect to the $i$-th person.
Thus, the weighted cross-entropy loss is written as
\begin{equation}\label{eq07}
\begin{aligned}
&\mathcal{L}_\mathrm{WC}= -\sum_{c=1}^{C}W_{z}(c)\log P(c|x_i),
  \end{aligned}
\end{equation} where the sample $x_i$ belongs to the $z$-th person.

\emph{Remarks.} Compared with the soft classification in the literature~\cite{li2019unsupervised}, there are two main differences: i) Our work mainly focuses on the unsupervised cross-camera learning, while \cite{li2019unsupervised} aims at unsupervised within-camera learning; ii) Since our task is different from the task in \cite{li2019unsupervised}, to obtain better cross-camera soft-labels, we consider not only the $k$ nearest samples but also non-overlapping IDs in the same camera (i.e., neglecting the relationships between within-camera persons when computing the affinity matrix), whose effectiveness is validated by the experiments in Section~\ref{sec:EXP-FA}. 

 In addition, there are two discrepancies between label smooth~\cite{DBLP:conf/iccv/ZhengZY17} and our method. i) The ``equal label'' in label smooth is given for ``all classes''. However, our method assigns the soft-label for these classes whose value varies according to the affinity matrix of the cross-camera classes. Besides, for one image or tracklet, our method only assigns labels to the $k$ nearest classes with it, because one person (identity) usually includes a few classes but not all classes. ii) For label smooth, all labels are fixed at each iteration, while our method dynamically updates the affinity matrix and reassigns the better soft-label for each class at each iteration.

\subsubsection{Discrimination Task}
Besides utilizing the weighted cross-entropy loss to learn the feature representation with the soft-label information, we also employ the weighted triplet loss from the discrimination view to conduct unsupervised inter-camera person Re-ID. For this loss function, a key task is selecting appropriate positive and negative samples with the soft-label information to generate effective triplets, which is elaborated as follows. 

 \textbf{Selecting positive samples.} 
 We select positive samples according to the generated soft-labels. In this work, for an anchor, we ``randomly'' select the $N_K$ persons from the $k$ nearest neighbors. It is worth noting that the efficacy of random sampling has been verified in Section~\ref{sec:EXP-FA}, which could choose some underlying hard positive samples when compared with taking $N_K$ nearest persons with the anchor. For each person, we randomly select an image as a positive sample. Thus, we have $N_K$ positive samples for an anchor, which is consistent with the image number of each person in the supervised within-camera learning. Concretely, for an anchor $v_a$, we can get a positive sample set $\mathrm{P}_a$ and corresponding weight set $\mathrm{W}_a$ (i.e., each element in $\mathrm{W}_a$ denotes the similarity between the anchor and each its positive sample) from the above affinity matrix. For the weight values of positive samples, we normalize them to guarantee $\sum_{w_p\in \mathrm{W}_a} w_p = 1$.

\textbf{Selecting negative samples.} 
For negative samples, we use the same scheme with the hard sample mining to select the hardest negative samples from ``the same camera'' with the anchor because we can guarantee that the selected negative samples are true negative samples. Thus, for an anchor $v_a$, we can get the hardest negative sample $v_{hn}$ by the above method.


\textbf{Weighted triplet loss.} 
According to the positive and negative samples, we can formulate a weighted triplet loss as
 \begin{equation}\label{eq09}
\begin{aligned}
\mathcal{L}_{\mathrm{WT}}=\left [\sum_{(v_p, w_p)\in [\mathrm{P}_a, \mathrm{W}_a]} w_{p}D(v_a, v_p) -D(v_a, v_{hn}) + m \right ]_{+}.
\end{aligned}
\end{equation}

\emph{Remarks.} Compared with the weighted triplet loss in \cite{Ristani_2018_CVPR}, there are two main differences: i) Our loss function (Eq.~(\ref{eq09})) is used to solve the unsupervised task, while the loss in \cite{Ristani_2018_CVPR} is designed to address the supervised task. ii) Due to the property of our task (i.e., we can guarantee that the true negative samples are chosen by the above selecting negative sample method), instead of using a weighted scheme for both positive and negative samples in the literature \cite{Ristani_2018_CVPR}, we only utilize the weighted scheme for positive samples. 

Besides, through extensive experiments, we observe that using the average weighted scheme could get better results in our unsupervised task, i.e., all $w_i\in \mathrm{W}_a$ is set as $\frac{1}{N_K}$. The main reason is that the larger weight (i.e., the similarity between one person and the anchor person) corresponds to the easier positive sample. As analyzed in the literature~\cite{hermans2017defense}, using hard positive samples can bring better results. Therefore, the average weighted scheme could pay more attention to some hard positive samples. In Section~\ref{sec:EXP-FA}, we have conducted experiments to confirm the above observation.

\subsection{Optimization}\label{SEC:OPT}

The whole loss function of the proposed network is expressed as 
\begin{equation}\label{eq10}
\begin{aligned}
&\mathcal{L}_{\mathrm{Total}}= \mathcal{L}_{\mathrm{Intra}}+\lambda \mathcal{L}_{\mathrm{Inter}},\\
&\mathcal{L}_{\mathrm{Inter}} \in \left\{\mathcal{L}_{\mathrm{WC}}, \mathcal{L}_{\mathrm{WT}}\right\},
\end{aligned}
\end{equation} where $\lambda$ is the parameter to trade off the supervised intra-camera task and the unsupervised inter-camera task.

In the training stage, we first train the $T^{'}$ epochs with the intra-camera labels and thus get better feature representation to generate the affinity matrix. Then we jointly conduct the supervised intra-camera task and the unsupervised inter-camera task to update the parameters in the network. The whole training process is described in Algorithm~\ref{alg1}.

\begin{algorithm}[t]
\caption{Progressive Cross-camera Soft-label Learning (PCSL) for semi-supervised person Re-ID}~\label{alg1}
\begin{algorithmic}[1]
\STATE {\bf procedure} 
PCSL (Training examples $\mathrm{X}$, labels $\mathrm{Y}_{Intra}$) \\
\STATE $\theta \leftarrow$ Initialize by ResNet-50 pre-trained on ImageNet.\\
\FOR{epoch $\in [1,...,T^{'}]$}
\STATE $\theta \leftarrow$ OPTIMIZE($\mathcal{L}_{\mathrm{Intra}}$ in Eq.~(\ref{eq01})).
\STATE $\mathbf{P} \leftarrow$ Update by Eq.~(\ref{eq03}).
\ENDFOR
\FOR{epoch $\in [T^{'},...,T]$}
\STATE Compute the affinity matrix $\mathbf{A}$ by Eq.~(\ref{eq04}).
\STATE Generate cross-camera soft-labels.
\STATE $\theta \leftarrow$ OPTIMIZE($\mathcal{L}_{\mathrm{Total}}$ in Eq.~(\ref{eq10})).
\STATE $\mathbf{P} \leftarrow$ Update by Eq.~(\ref{eq03}).
\ENDFOR
\STATE {\bf end procedure}
\end{algorithmic}
\end{algorithm}
\section{Experiments}\label{s-experiment}
\renewcommand{\cmidrulesep}{0mm} 
\setlength{\aboverulesep}{0mm} 
\setlength{\belowrulesep}{0mm} 
\setlength{\abovetopsep}{0cm}  
\setlength{\belowbottomsep}{0cm}
\begin{table*}[htbp]
  \centering
  \caption{Comparison with the state-of-the-art unsupervised methods on three image datasets including Market1501, DukeMTMC-reID and MSMT17. ``-'' denotes that the result is not provided. mAP and the Rank-1, 5, 10 accuracies of CMC are reported in this table. The best performance is \textbf{bold}.}
    \begin{tabular}{|c|cccc|cccc|cccc|}
    \toprule
    \midrule
    \multirow{2}[1]{*}{Method} & \multicolumn{4}{c|}{Market1501} & \multicolumn{4}{c|}{DukeMTMC-reID} & \multicolumn{4}{c|}{MSMT17} \\
\cmidrule{2-13}          & mAP   & Rank-1 & Rank-5 & Rank-10 & mAP   & Rank-1 & Rank-5 & Rank-10 & mAP   & Rank-1 & Rank-5 & Rank-10 \\
    \midrule
    LOMO~\cite{liao2015person}  & 8.0   & 27.2  & 41.6  & 49.1  & 4.8   & 12.3  & 21.3  & 26.6  & -    & -    & -    & - \\
    BoW~\cite{DBLP:conf/iccv/ZhengSTWWT15}   & 14.8  & 35.8  & 52.4  & 60.3  & 8.3   & 17.1  & 28.8  & 34.9  & -    & -    & -    & - \\
    UJSDL~\cite{qi2018unsupervised} & -    & 50.9  & -    & -    &    -   & 32.2  & -    & -    & -    & -    & -    & - \\
    UMDL~\cite{DBLP:conf/cvpr/PengXWPGHT16}  & 12.4  & 34.5  & -    & -    & 7.3   & 18.5  & -    & -    & -    & -    & -    & - \\
    CAMEL~\cite{DBLP:conf/iccv/YuWZ17} & 26.3  & 54.5  & -    & -    &   -    &   -    & -    & -    & -    & -    & -    & - \\
    \midrule
    PUL~\cite{fan2018unsupervised}   & 20.5  & 45.5  & 60.7  & 66.7  & 16.4  & 30.0  & 43.4  & 48.5  & -    & -    & -    & - \\
    Tfusion~\cite{lv2018unsupervised} & -    & 60.8  & -    & -    & -    & -    & -    & -    & -    & -    & -    & - \\
    TJ-AIDL~\cite{wang2018transferable}  & 26.5  & 58.2  & 74.8  & 81.1  & 23.0  & 44.3  & 59.6  & 65.0  & -    & -    & -    & - \\
    MAR~\cite{Yu_2019_CVPR} & 40.0  & 67.7  & 81.9  & 87.3  & 48.0  & 67.1   & 79.8   & 84.2  & -    & -    & -    & - \\
    \midrule
    MMFA~\cite{DBLP:conf/bmvc/LinLLK18}  & 27.4  & 56.7  & 75.0  & 81.8  & 24.7  & 45.3  & 59.8  & 66.3  & -    & -    & -    & - \\
    CAT~\cite{DBLP:journals/corr/abs-1904-01308}   & 27.8  & 57.8  & -    & -    & 28.7  & 50.9  & -    & -    & -    & -    & -    & - \\
    CAL-CCE~\cite{qi2019novel} & 34.5  & 64.3  & -    & -    & 36.7  & 55.4  & -    & -    & -    & -    & -    & - \\
    \midrule
    PTGAN~\cite{wei2018person} & -    & 38.6  & -    & -    & -    & 27.2  & -    & -    & 3.3   & 11.8  & -    & 27.4 \\
    SPGAN~\cite{deng2018image} & 22.8  & 51.5  & 70.1  & 76.8  & 22.3  & 41.1  & 56.6  & 63.0  & -    & -    & -    & - \\
    SPGAN+LMP~\cite{deng2018image}  & 26.7  & 57.7  & 75.8  & 82.4  & 26.2  & 46.4  & 62.3  & 68.0  & -    & -    & -    & - \\
    HHL~\cite{zhong2018generalizing}   & 31.4  & 62.2  & 78.8  & 84.0  & 27.2  & 46.9  & 61.0  & 66.7  & -    & -    & -    & - \\
    UTA~\cite{tian2019imitating}   & 40.1  & 72.4  & 87.4  & 91.4  & 31.8  & 55.6  & 68.3  & 72.4  & -    & -    & -    & - \\
    ECN~\cite{zhong2019invariance}   & 43.0  & 75.1 & 87.6 & 91.6  & 40.4  & 63.3  & 75.8  & 80.4  & 10.2  & 30.2  & 41.5  & 46.8 \\
    \midrule
    PCSL-C  (ours) & \textbf{69.4} & \textbf{87.0} & \textbf{94.8} & \textbf{96.6} & 50.0  & 70.6 & 82.4  & 85.7  & \textbf{20.7} & \textbf{48.3} & \textbf{62.8} & \textbf{68.6} \\
    PCSL-D (ours) & 65.4  & 82.0  & 92.2  & 95.1  & \textbf{53.5} & \textbf{71.7} & \textbf{84.7} & \textbf{88.2} & 20.5  & 45.1  & 61.0  & 67.6 \\
    \bottomrule
    \end{tabular}%
  \label{tab01}%
  \vspace*{-10pt}%
\end{table*}%

In this part, we firstly introduce the experimental datasets and settings in Section~\ref{sec:EXP-DS}. Then, we compare the proposed method with the state-of-the-art unsupervised Re-ID methods and some methods with the semi-supervised setting in Sections~\ref{sec:EXP-CUA} and~\ref{sec:EXP-SS}, respectively. Furthermore, we compare our proposed method with deep supervised person Re-ID methods in Section~\ref{sec:EXP-CSM}. To validate the effectiveness of various components in the proposed framework, we conduct ablation studies in Section~\ref{sec:EXP-EDC}. Lastly, we further analyze the property of the proposed network in Section~\ref{sec:EXP-FA}.
\subsection{Datasets and Experiment Settings}\label{sec:EXP-DS}
We evaluate our method on three large-scale image datasets: Market1501~\cite{DBLP:conf/iccv/ZhengSTWWT15}, DukeMTMC-reID~\cite{DBLP:conf/iccv/ZhengZY17}, and MSMT17~\cite{wei2018person}. 
 \textbf{Market1501} contains 1,501 persons with 32,668 images from six cameras. Among them, $12,936$ images of $751$ identities are used as training set. For evaluation, there are $3,368$ and $19,732$ images in the query set and the gallery set, respectively. \textbf{DukeMTMC-reID} has $1,404$ persons from eight cameras, with $16,522$ training images, $2,228$ queries, and $17,661$ gallery images.
 \textbf{MSMT17} is collected from a 15-camera network deployed on campus. The training set contains $32,621$ images of $1,041$ identities. For evaluation, $11,659$ and $82,161$ images are adopted as query and gallery images, respectively. For all datasets, we employ CMC accuracy and mAP for Re-ID evaluation~\cite{DBLP:conf/iccv/ZhengSTWWT15}. On Market1501, there are single- and multi-query evaluation protocols. We use the more challenging single-query protocol in our experiments.

In addition, we also evaluate the proposed method on two large-scale video datasets including MARS~\cite{DBLP:conf/eccv/ZhengBSWSWT16} and DukeMTMC-SI-Tracklet~\cite{li2019unsupervised}. \textbf{MARS} has a total of $20,478$ tracklets of $1,261$ persons captured from a 6-camera network at a university campus. All the tracklets were automatically generated by the DPM detector~\cite{felzenszwalb2009object} and the GMMCP tracker~\cite{dehghan2015gmmcp}. This dataset splits $626$ and $635$ identities into training and testing sets, respectively. \textbf{DukeMTMC-SI-Tracklet} is from DukeMTMC. It consists of $19,135$ person tracklets and $1,788$ persons from $8$ cameras. In this dataset, $702$ and $1,086$ identities are dived into training and testing sets, respectively. On these video datasets, we also employ CMC accuracy and mAP for Re-ID evaluation~\cite{DBLP:conf/iccv/ZhengSTWWT15}.

To conduct the classification task with the weighted cross-entropy loss, we randomly select the same number (i.e., $\left \lfloor \frac{64}{N_C} \right \rfloor$, where $N_C$ is the number of cameras and $\left \lfloor \cdot  \right \rfloor$ denotes the round down operation) of images per camera in a batch. In the discrimination task, we set $N_P$ (i.e., the number of persons) and $N_K$ (i.e., the number of images per person) as $32$ and $4$ to produce triplets, which is the same with the literature~\cite{hermans2017defense}. The margin of triplet loss, $m$, is $0.3$ according to the literature~\cite{hermans2017defense}. $\lambda$ in Eq. (\ref{eq10}) is set as $1$. When generating affinity matrix, we only consider $6$ (i.e., $k=6$) nearest persons.
 The initial learning rates of the fine-tuned parameters (those in the pre-trained ResNet-50 on ImageNet~\cite{DBLP:conf/cvpr/DengDSLL009}) and the new parameters (those in the newly added layers) are $0.1$ and $0.01$, respectively. The proposed model is trained with the SGD optimizer in a total of $300$ epochs (i.e., $T=300$). $T'$ is set as $100$. When the number of epochs at $200$, we decrease the learning rates by a factor of $0.1$. The size of the input image is $256 \times 128$. Particularly, all experiments on all datasets utilize the same experimental settings. In the testing stage, we do not utilize the camera information or any other extra information. 
 For each testing image, we only normalize the raw image by the unified mean value and standard deviation, which is commonly used in all the existing deep methods for person Re-ID. In our paper, for a fair comparison, we only report the result of the single-query evaluation protocol on all datasets.
  In all experimental results, PCSL-C and PCSL-D denote the inter-camera learning based on classification (C) and discrimination (D) views, respectively.

\subsection{Comparison with Unsupervised Methods}\label{sec:EXP-CUA}
We compare our method with the state-of-the-art unsupervised image-based person Re-ID approaches. 
Among them, there are five non-deep-learning-based methods (LOMO~\cite{liao2015person}, BoW~\cite{DBLP:conf/iccv/ZhengSTWWT15}, UJSDL~\cite{qi2018unsupervised}, UMDL~\cite{DBLP:conf/cvpr/PengXWPGHT16} and CAMEL~\cite{DBLP:conf/iccv/YuWZ17}), and multiple deep-learning-based methods. The latter includes four recent pseudo-label-generation-based methods (PUL~\cite{fan2018unsupervised}, TFusion~\cite{lv2018unsupervised}, TJ-AIDL~\cite{wang2018transferable} and MAR~\cite{Yu_2019_CVPR}), three distribution-alignment-based approaches (MMFA~\cite{DBLP:conf/bmvc/LinLLK18}, CAT~\cite{DBLP:journals/corr/abs-1904-01308} and CAL-CCE~\cite{qi2019novel}) and five recent image-generation-based approaches (PTGAN~\cite{wei2018person}, SPGAN~\cite{deng2018image}, HHL~\cite{zhong2018generalizing},  UTA~\cite{tian2019imitating} and ECN~\cite{zhong2019invariance}). The experimental results on Market1501, DukeMTMC-reID and MSMT17 are reported in Table~\ref{tab01}. As seen, our approach (i.e., PCSL-C and PCSL-D) consistently shows the superiority over all compared methods on all datasets. For example, the proposed method significantly outperforms the recent pseudo-label-generation-based methods, such as PUL and CAMEL. This contributes to the available intra-camera label information and progressive cross-camera soft-label learning. Particularly, compared with recent ECN~\cite{zhong2018generalizing}, the state-of-the-art by utilizing both labeled source domain and generated images from the GAN-based model, PCSL-C gains $26.4\%$ ($69.4$ vs. $43.0$), $9.6\%$ ($50.0$ vs. $40.4$) and $10.5\%$ ($20.7$ vs. $10.2$) in mAP on Market1501, DukeMTMC-reID and MSMT17, respectively. Particularly, the most existing unsupervised methods utilize labeled source domains and generated images by the GAN-based methods. Differently, our method does not use the other extra data.

Moreover, we also compare our method with the state-of-the-art unsupervised video-based Re-ID approaches on MARS and DukeMTMC-SI-Tracklet, which include
GRDL~\cite{kodirov2016person}, UnKISS~\cite{khan2016unsupervised}, RACE~\cite{ye2018robust}, Stepwise~\cite{liu2017stepwise}, DGM+MLAPG~\cite{ye2017dynamic}, DGM+IDE~\cite{ye2017dynamic}, DAL~\cite{DBLP:conf/bmvc/ChenZG18}, TAUDL~\cite{Li_2018_ECCV} and UTAL~\cite{li2019unsupervised}. Among them, except for GRDL and UnKISS, all other methods are based on deep features. The results are reported in Table~\ref{tab02}. Compared with recent UTAL, which has shown the superiority of the proposed method in the unsupervised video-based Re-ID task, our semi-supervised methods (PCSL-C and PCSL-D) achieve a significant improvement in mAP and CMC accuracy. In particular, PCSL-C increases UTAL by $27.0\%$ ($62.2$ vs. $35.2$) and $20.6\%$ ($70.5$ vs. $49.9$) in mAP and Rank-1 accuracy on MARS, respectively. This further validates the effectiveness of the proposed methods.

\begin{table}
  \centering
  \caption{Comparison with the state-of-the-art unsupervised methods on two video datasets including MARS and DukeMTMC-SI-Tracklet (Duke-T). We report mAP and the Rank-1, 5, 20 accuracies of CMC in this table.}
    \begin{tabular}{|c|c|cccc|}
    \toprule
    \midrule
          & Method & mAP   & Rank-1 & Rank-5 & Rank-20 \\
          \midrule
    \multirow{11}[1]{*}{\begin{sideways}MARS\end{sideways}} & GRDL~\cite{kodirov2016person}  & 9.6   & 19.3  & 33.2  & 46.5 \\
          & UnKISS~\cite{khan2016unsupervised} & 10.6  & 22.3  & 37.4  & 53.6 \\
          & RACE~\cite{ye2018robust}  & 24.5  & 43.2  & 57.1  & 67.6 \\
          & Stepwise~\cite{liu2017stepwise} & 10.5  & 23.6  & 35.8  & 44.9 \\
          & DGM+MLAPG~\cite{ye2017dynamic} & 11.8  & 24.6  & 42.6  & 57.2 \\
          & DGM+IDE~\cite{ye2017dynamic} & 21.3  & 36.8  & 54.0  & 68.5 \\
          & DAL~\cite{DBLP:conf/bmvc/ChenZG18}   & 21.4  & 46.8  & 63.9  & 77.5 \\
          & TAUDL~\cite{Li_2018_ECCV} & 29.1  & 43.8  & 59.9  & 72.8 \\
          & UTAL~\cite{li2019unsupervised}  & 35.2  & 49.9  & 66.4  & 77.8 \\
\cmidrule{2-6}           & PCSL-C  (ours) & \textbf{62.2} & \textbf{70.5} & \textbf{87.0} & \textbf{94.2} \\
          & PCSL-D (ours) & 57.5  & 65.6  & 83.1  & 91.9 \\
    \midrule
    \midrule
    \multirow{4}[1]{*}{\begin{sideways}Duke-T\end{sideways}} & TAUDL~\cite{Li_2018_ECCV} & 20.8  & 26.1  & 42.0  & 57.2 \\
          & UTAL~\cite{li2019unsupervised}  & 36.6  & 43.8  & 62.8  & 76.5 \\
\cmidrule{2-6}         & PCSL-C  (ours) & 37.3  & 47.3  & 65.4  & 75.9 \\
          & PCSL-D (ours) & \textbf{43.1} & \textbf{52.3} & \textbf{72.3} & \textbf{84.4} \\
    \bottomrule
    \end{tabular}%
  \label{tab02}%
   \vspace{-15pt}%
\end{table}%

\subsection{Comparison with Methods in Semi-supervised Setting} \label{sec:EXP-SS}
In this section, we compare our method with the unsupervised video-based person Re-ID methods in the semi-supervised setting (i.e., provide the intra-camera labels but not inter-camera labels). Both TAUDL~\cite{Li_2018_ECCV} and UTAL~\cite{li2019unsupervised} consist of intra-camera tracklet discrimination learning and cross-camera tracklet association learning. In the semi-supervised setting, they construct the connection of different cameras by self-discovering the cross-camera positive matching pairs. Different from both TAUDL~\cite{Li_2018_ECCV} and UTAL~\cite{li2019unsupervised}, we conduct the unsupervised cross-camera learning by assigning progressive soft-labels. We show the experimental results in Tables~\ref{tab03} and~\ref{tab04}. As seen, on all image datasets (Table~\ref{tab03}), the proposed method has excellent performance compared with TAUDL and UTAL. For example, the Rank-1 accuracy of PCSP-C gains $17.8\%$ ($87.0$ vs. $69.2$), $8.3\%$ ($70.6$ vs. $62.3$) and $16.9\%$ ($48.3$ vs. $31.4$) over UTAL on Market1501, DukeMTMC-reID and MSMT17, respectively. Besides, on video datasets, our method can also achieve good performance. PCSL-D can improve $5.8\%$ ($57.5$ vs. $51.7$) and $3.9\%$ ($43.1$ vs. $39.0$) in mAP over UTAL on MARS and DukeMTMC-SI-Tracklet, respectively.

Besides, we also compare our method with the pseudo-label-based method with one-hot label~\cite{lee2013pseudo}, which is designed to solve the semi-supervised classification task. The experimental results are shown in Table~\ref{tab03}. As seen, the method in~\cite{lee2013pseudo} cannot work well for our semi-supervised person Re-ID task, because our task is different from the classification task in the semi-supervised setting. Concretely, a semi-supervised classification task gives the total number of categories (i.e., persons) on a dataset and has few labeled samples for each category. However, in our task, we do not know the total number of persons (i.e., categories) on a dataset and only provides labels for the within-camera images. Therefore, different from the methods for the typical semi-supervised classification task, our goal is to explore the relationship between cross-camera persons. Particularly, since a person could appear in multiple different cameras, if we only use the one-hot label, it will imply that the person only appears in two different cameras. 
As a result, the method in~\cite{lee2013pseudo} cannot obtain good performance.
 \vspace*{-15pt}%
\newcommand{\PreserveBackslash}[1]{\let\temp=\\#1\let\\=\temp}
\newcolumntype{C}[1]{>{\PreserveBackslash\centering}p{#1}}
\newcolumntype{R}[1]{>{\PreserveBackslash\raggedleft}p{#1}}
\newcolumntype{L}[1]{>{\PreserveBackslash\raggedright}p{#1}}
\begin{table}
  \centering
  \caption{Comparison with TAUDL and UTAL in the semi-supervised setting on Market1501, DukeMTMC-reID (Duke) and MSMT17. The best performance is \textbf{bold}.}
    \begin{tabular}{|C{2.2cm}|C{0.3cm}C{0.9cm}|C{0.3cm}C{0.9cm}|C{0.3cm}C{0.9cm}|}
    \toprule
    \midrule
    \multirow{2}[1]{*}{Method} & \multicolumn{2}{c|}{Market1501} & \multicolumn{2}{c|}{Duke} & \multicolumn{2}{c|}{MSMT17} \\
\cmidrule{2-7}          & mAP   & Rank-1 & mAP   & Rank-1 & mAP & Rank-1 \\
    \midrule
    One-hot~\cite{lee2013pseudo} & 24.6  & 48.5  & 30.6  & 43.9  & 10.3 & 24.2 \\
    TAUDL~\cite{Li_2018_ECCV} & 41.2  & 63.7  & 43.5  & 61.7  & 12.5 & 28.4 \\
    UTAL~\cite{li2019unsupervised}  & 46.2  & 69.2  & 44.6  & 62.3  &  13.1 & 31.4 \\
    \midrule
    PCSL-C  (ours) & \textbf{69.4} & \textbf{87.0} & 50.0  & 70.6 & \textbf{20.7} & \textbf{48.3} \\
    PCSL-D (ours) & 65.4  & 82.0  & \textbf{53.5} & \textbf{71.7}  & 20.5  & 45.1 \\
    \bottomrule
    \end{tabular}%
  \label{tab03}%
\end{table}%
\begin{table}
  \centering
  \caption{Comparison with UTAL in the semi-supervised setting on MARS and DukeMTMC-SI-Tracklet (Duke-T).}
    \begin{tabular}{|c|cc|cc|}
    \toprule
    \midrule
    \multirow{2}[1]{*}{Method } & \multicolumn{2}{c|}{MARS} & \multicolumn{2}{c|}{Duke-T} \\
\cmidrule{2-5}          & mAP   & Rank-1 & mAP   & Rank-1 \\
    \midrule
    UTAL~\cite{li2019unsupervised}  & 51.7 & 59.5 & 39.0  & 46.4 \\
    \midrule
    PCSL-C  (ours) & \textbf{62.2} & \textbf{70.5} & 37.3  & 47.3 \\
    PCSL-D (ours) & 57.5  & 65.6  & \textbf{43.1} & \textbf{52.3} \\
    \bottomrule
    \end{tabular}%
  \label{tab04}%
   \vspace*{-15pt}%
\end{table}%
\subsection{Comparison with Supervised Methods}\label{sec:EXP-CSM}
In this part, we also compare the proposed method and several supervised methods on Market1501. The compared methods include three non-deep-learning-based methods (LOMO + XQDA~\cite{DBLP:conf/cvpr/LiaoHZL15},  BoW+Kissme~\cite{DBLP:conf/iccv/ZhengSTWWT15} and DNSL~\cite{DBLP:conf/cvpr/ZhangXG16}) and thirteen deep-learning-based methods (PersonNet~\cite{wu2016personnet}, Gate S-CNN~\cite{DBLP:conf/eccv/VariorHW16}, LSTM S-CNN~\cite{DBLP:conf/eccv/VariorSLXW16}, DGDropout~\cite{DBLP:conf/cvpr/XiaoLOW16}, Deep-Embed~\cite{wu2018deep}, SpindleNet~\cite{zhao2017spindle}, Part-Aligned~\cite{Zhao_2017_ICCV}, PIE~\cite{zheng2019pose}, JLML~\cite{liu2017stepwise}, MTMCT~\cite{Ristani_2018_CVPR}, SVDNet~\cite{sun2017svdnet}, PDC~\cite{su2017pose} and A$^3$M~\cite{han2018attribute}).  Particularly, PIE, MTMCT, SVDNet and A$^3$M use the same backbone network (i.e., ResNet-50~\cite{DBLP:conf/cvpr/HeZRS16}) with our method. The experimental results are reported in Table~\ref{tab06}. As seen, our proposed method outperforms most compared supervised methods. For example, PCSL-C significantly increases $34.7\%$ ($69.4$ vs. $35.7$) and $26\%$ ($87.0$ vs. $61.0$) in mAP and Rank-1 accuracy over the best non-deep-learning DNSL. For the recent deep-learning method A$^3$M, which
can learn local attribute representation and global category representation simultaneously in an end-to-end manner, our method still obtain a competitive performance. Different from the supervised methods that need expensive label information, especially for the inter-camera labels, the proposed method only need the intra-camera labels which can be readily captured by employing tracking algorithms and conducting few manual annotations. Therefore, these comparisons further show the proposed semi-supervised method is very meaningful in the person Re-ID community.
\begin{table}[htbp]
  \centering
  \caption{Comparison with supervised methods on Market1501. ``-'' denotes that the result is not provided. The best performance is \textbf{bold}}
    \begin{tabular}{|c|cccc|}
    \toprule
    \midrule
    Method & mAP   & Rank-1 & Rank-5 & Rank-10 \\
    \midrule
    LOMO + XQDA~\cite{DBLP:conf/cvpr/LiaoHZL15} & 22.2  & 43.8  &  -     & - \\
    BoW+Kissme~\cite{DBLP:conf/iccv/ZhengSTWWT15} & 20.8  & 44.4  & 63.9  & 72.2 \\
    DNSL~\cite{DBLP:conf/cvpr/ZhangXG16}  & 35.7  & 61.0  &  -     & - \\
    \midrule
    PersonNet~\cite{wu2016personnet} & 26.4  & 37.2  &  -     & - \\
    Gate S-CNN~\cite{DBLP:conf/eccv/VariorHW16} & 39.6  & 65.9  &  -     & - \\
    LSTM S-CNN~\cite{DBLP:conf/eccv/VariorSLXW16}  & 35.3  & 61.6  & -      & - \\
    DGDropout~\cite{DBLP:conf/cvpr/XiaoLOW16}  & 31.9  & 59.5  &  -     & - \\
    Deep-Embed~\cite{wu2018deep} & 40.2  & 68.3  & 87.2  & 94.6 \\
    SpindleNet~\cite{zhao2017spindle} &   -    & 76.9  & 91.5  & 94.6 \\
    Part-Aligned~\cite{Zhao_2017_ICCV} &   -    & 81.0  & 92.0  & 94.7 \\
    MSCAN~\cite{li2017learning} & 57.5  & 80.3  & 92.0  & - \\
    PIE~\cite{zheng2019pose}   & 53.9  & 79.3  & 90.7  & 94.4 \\
    JLML~\cite{liu2017stepwise}  & 65.5  & 85.1  &   -    & - \\
    MTMCT~\cite{Ristani_2018_CVPR} & 68.0  & 84.2  &  -     & - \\
    SVDNet~\cite{sun2017svdnet} & 62.1  & 82.3  & 92.3  & 95.2 \\
    PDC~\cite{su2017pose}   & 63.4  & 84.1  & 92.7  & 94.9 \\
    A$^3$M~\cite{han2018attribute}   & 69.0  & 86.5  & \textbf{95.2}  & \textbf{97.0} \\
    \midrule
    PCSL-C  (ours) & \textbf{69.4} & \textbf{87.0} & 94.8 & 96.6 \\
    PCSL-D (ours) & 65.4  & 82.0  & 92.2  & 95.1 \\
    \bottomrule
    \end{tabular}%
  \label{tab06}%
    \vspace{-15pt}%
\end{table}%
\subsection{Effectiveness of Different Components in PCSL}\label{sec:EXP-EDC}
To sufficiently validate the efficacy of different components in the proposed network, we conduct experiments on five large-scale datasets. The experimental results are reported in Table \ref{tab05}. Note that ``only $\mathcal{L}_{\mathrm{Intra}}$'' denotes that merely conducting the intra-camera supervised task (i.e., in Eq.~(\ref{eq10}), $\lambda$ is set as $0$) in the proposed network.  Firstly, on all datasets, using progressive cross-camera soft-label learning can significantly improve the performance of the model with only supervised intra-camera learning (i.e., ``only $\mathcal{L}_{\mathrm{Intra}}$'' in Table \ref{tab05}). For example, PCSL-D can improve $23.2\%$ ($57.5$ vs. $34.3$) and $12.4\%$ ($43.1$ vs. $30.7$) in mAP on MARS and DukeMTMC-SI-Tracklet. Also, the Rank-1 accuracy can be improved by $20.1\%$ ($65.6$ vs. $45.5$) and $13.9\%$ ($52.3$ vs. $38.4$). This sufficiently validates the efficacy of the progressive cross-camera soft-label learning method.
Secondly, compared with results of PUL~\cite{fan2018unsupervised}, TFusion~\cite{lv2018unsupervised} and TJ-AIDL~\cite{wang2018transferable} in Table~\ref{tab01}, which are based on pseudo-label-generation, ``only $\mathcal{L}_{\mathrm{Intra}}$'' still has the competitive performance. This confirms using intra-camera labels indeed brings improvement. For example, ``only $\mathcal{L}_{\mathrm{Intra}}$'' improves $7.9\%$ ($34.4$ vs. $26.5$) and $18.7\%$ ($41.7$ vs. $23.0$) over TJ-AIDL in mAP on Market1501 and Duke.

 Lastly, PCSL-C outperforms PCSL-D on Market1501, MSMT17 and MARS, while the former is poorer than the latter on DukeMTMC-reID (Duke) and DukeMTMC-SI-Tracklet (Duke-T). The main reason is that the inter-camera learning on Duke and Duke-T is harder than other datasets. In this case, the discrimination task with cross-camera soft-labels can get better results. For example, on the Rank-1 accuracy, using the proposed unsupervised cross-camera learning can increase over $20\%$ on all datasets, except for Duke and Duke-T as shown in Table \ref{tab05}. This confirms that the cross-camera task is more difficult on both Duke and Duke-T when compared with other datasets. Particularly, since Duke and Duke-T are collected from the same video set, there is the same observation on two datasets. Therefore, in the case with the large variation across cameras, the discrimination method could be more appropriate to handle the semi-supervised task than the classification method. 
\begin{table}[htbp]
  \centering
  \caption{Evaluation of different components of PCSL on three image datasets (i.e., Market1501, DukeMTMC-reID and MSMT17) and two video datasets (i.e., DukeMTMC-SI-Tracklet (Duke-T) and MARS). The best performance is \textbf{bold}.}
    \begin{tabular}{|c|c|cccc|}
    \toprule
      \midrule
     Dataset & Method & mAP   & Rank-1 & Rank-5 & Rank-10\\
    \midrule
    \multirow{3}[1]{*}{Market1501} & Only $\mathcal{L}_{\mathrm{Intra}}$ & 34.4  & 58.1  & 72.5 & 78.0\\
\cmidrule{2-6}          & PCSL-C  (ours) & \textbf{69.4} & \textbf{87.0} & \textbf{94.8} & \textbf{96.6}\\
          & PCSL-D (ours) & 65.4  & 82.0  & 92.2 & 95.1\\
    \midrule
    \midrule
    \multirow{3}[1]{*}{Duke} & Only $\mathcal{L}_{\mathrm{Intra}}$ & 41.7  & 60.1  & 75.1 & 80.7 \\
\cmidrule{2-6}           & PCSL-C  (ours) & 50.0 & 70.6 & 82.4 & 85.7 \\
          & PCSL-D (ours) & \textbf{53.5} & \textbf{71.7} & \textbf{84.7} & \textbf{88.2} \\
    \midrule
    \midrule
    \multirow{3}[1]{*}{MSMT17} & Only $\mathcal{L}_{\mathrm{Intra}}$ & 10.0  & 24.8  & 38.1 & 44.8 \\
\cmidrule{2-6}          & PCSL-C  (ours) & \textbf{20.7} & \textbf{48.3} & \textbf{62.8} & \textbf{68.6}\\
          & PCSL-D (ours) & 20.5  & 45.1  & 61.0 & 67.6 \\
    \midrule
    \midrule
    \multirow{3}[1]{*}{MARS} & Only $\mathcal{L}_{\mathrm{Intra}}$ & 34.3  & 45.5  & 61.0 & 66.3 \\
\cmidrule{2-6}         & PCSL-C  (ours) & \textbf{62.2} & \textbf{70.5} & \textbf{87.0} & \textbf{90.7}\\
          & PCSL-D (ours) & 57.5  & 65.6  & 83.1 & 87.7 \\
    \midrule
    \midrule
    \multirow{3}[1]{*}{Duke-T} & Only $\mathcal{L}_{\mathrm{Intra}}$ & 30.7  & 38.4  & 55.5 & 62.6 \\
\cmidrule{2-6}           & PCSL-C  (ours) & 37.3  & 47.3  & 65.4 & 71.0 \\
          & PCSL-D (ours) & \textbf{43.1} & \textbf{52.3} & \textbf{72.3} & \textbf{78.3} \\
    \bottomrule
    \end{tabular}%
  \label{tab05}%
     \vspace{-15pt}%
\end{table}%

\subsection{Further Analysis}\label{sec:EXP-FA}

\textbf{Algorithm convergence.}
To investigate the convergence of our algorithm, we record the mAP and Rank-1 accuracy of PCSL-C and PCSL-D during the training on a validated set of DukeMTMC-reID in Fig.~\ref{fig6}. As seen, we can observe that our methods can almost converge after 200 epochs. 

\begin{figure}
\centering
\subfigure[PCSL-C]{
\includegraphics[width=4cm]{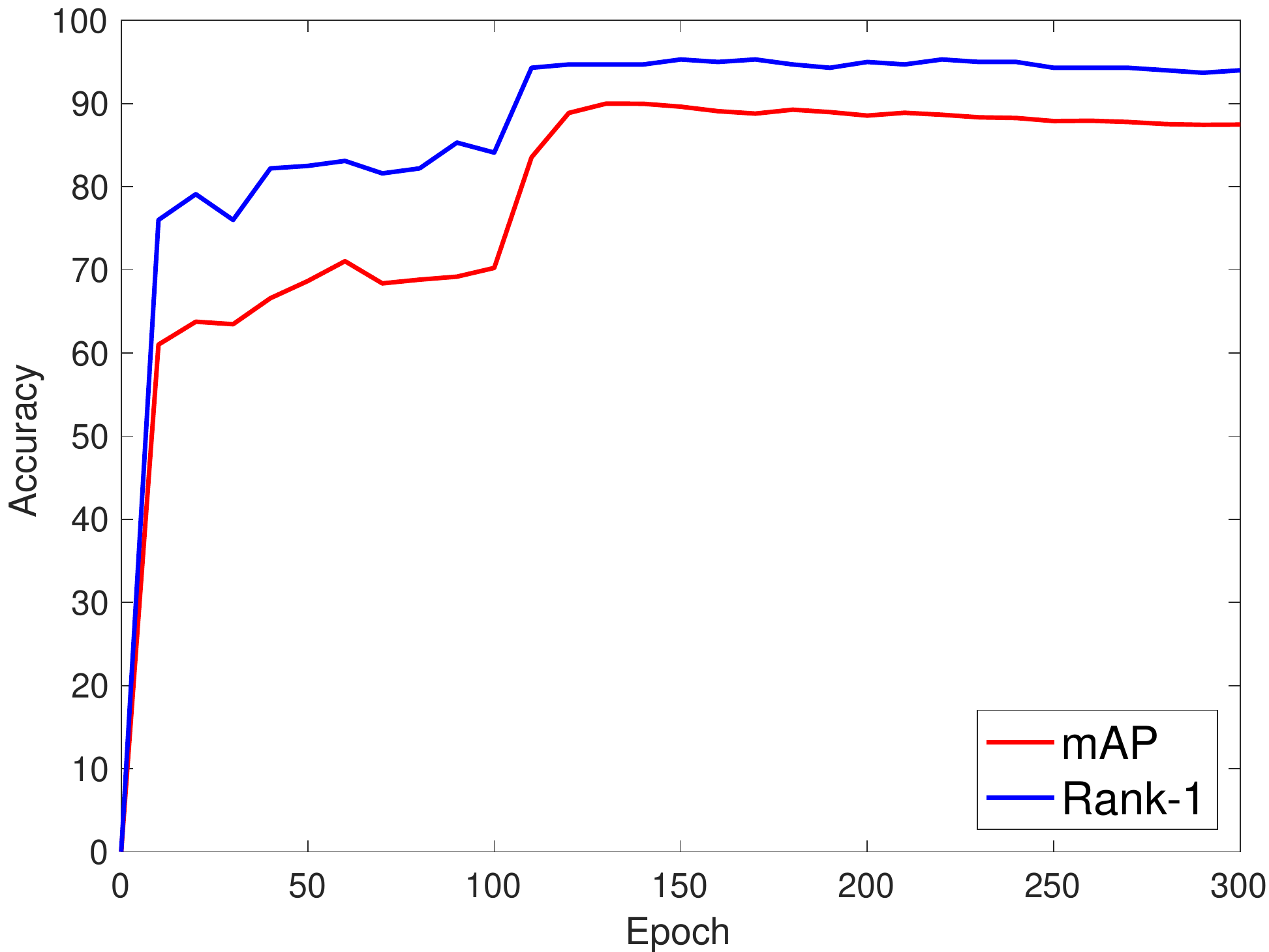}
}
\subfigure[PCSL-D]{
\includegraphics[width=4cm]{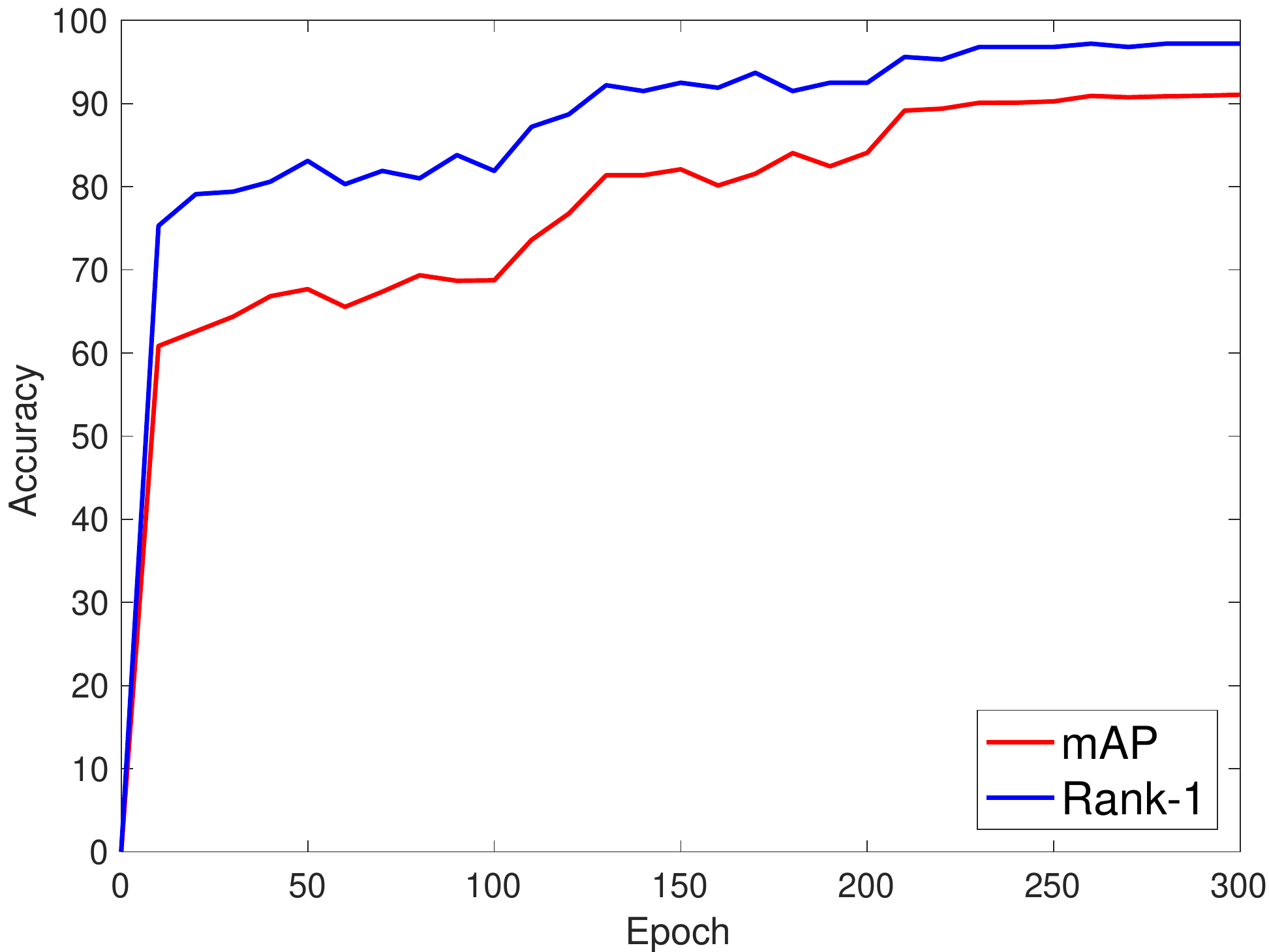}
}
\caption{Convergence curves of PCPL-C and PCPL-D on DukeMTMC-reID.}
\label{fig3}
   \vspace{-15pt}%
\end{figure}

\begin{figure}
\centering
\subfigure[$\lambda$]{
\includegraphics[width=4cm]{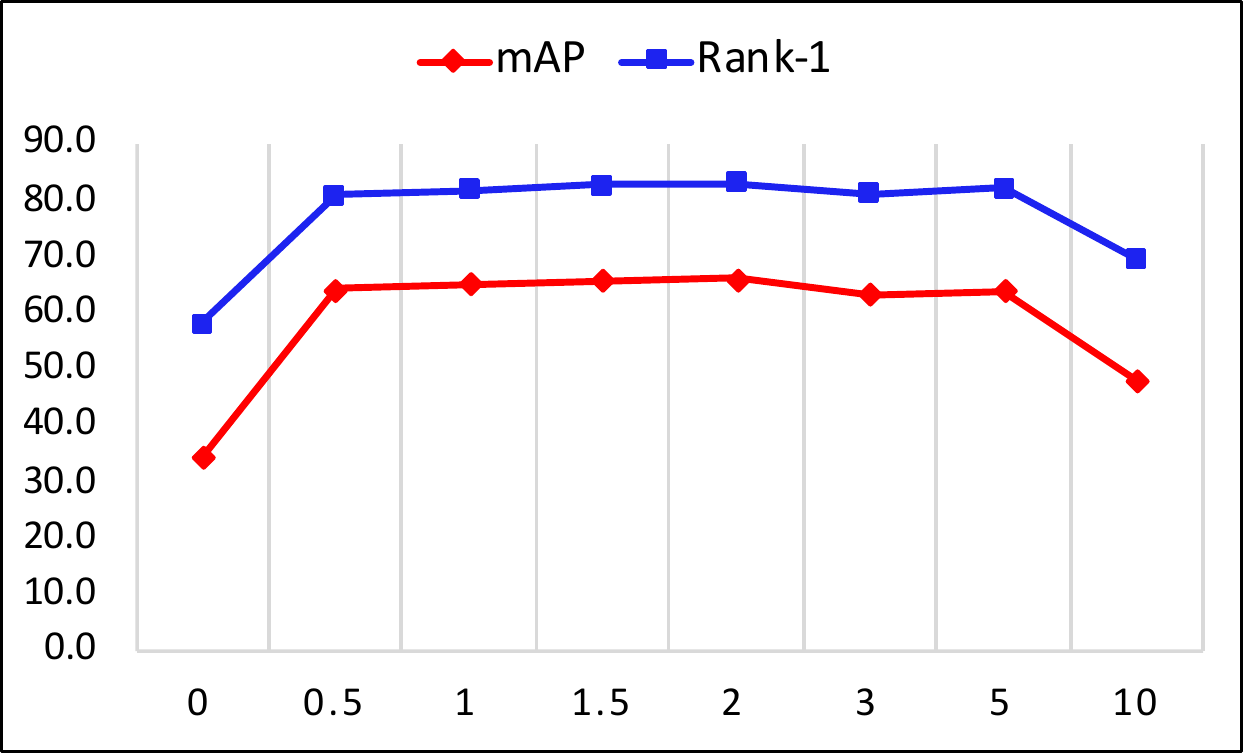}
}
\subfigure[$k$]{
\includegraphics[width=4cm]{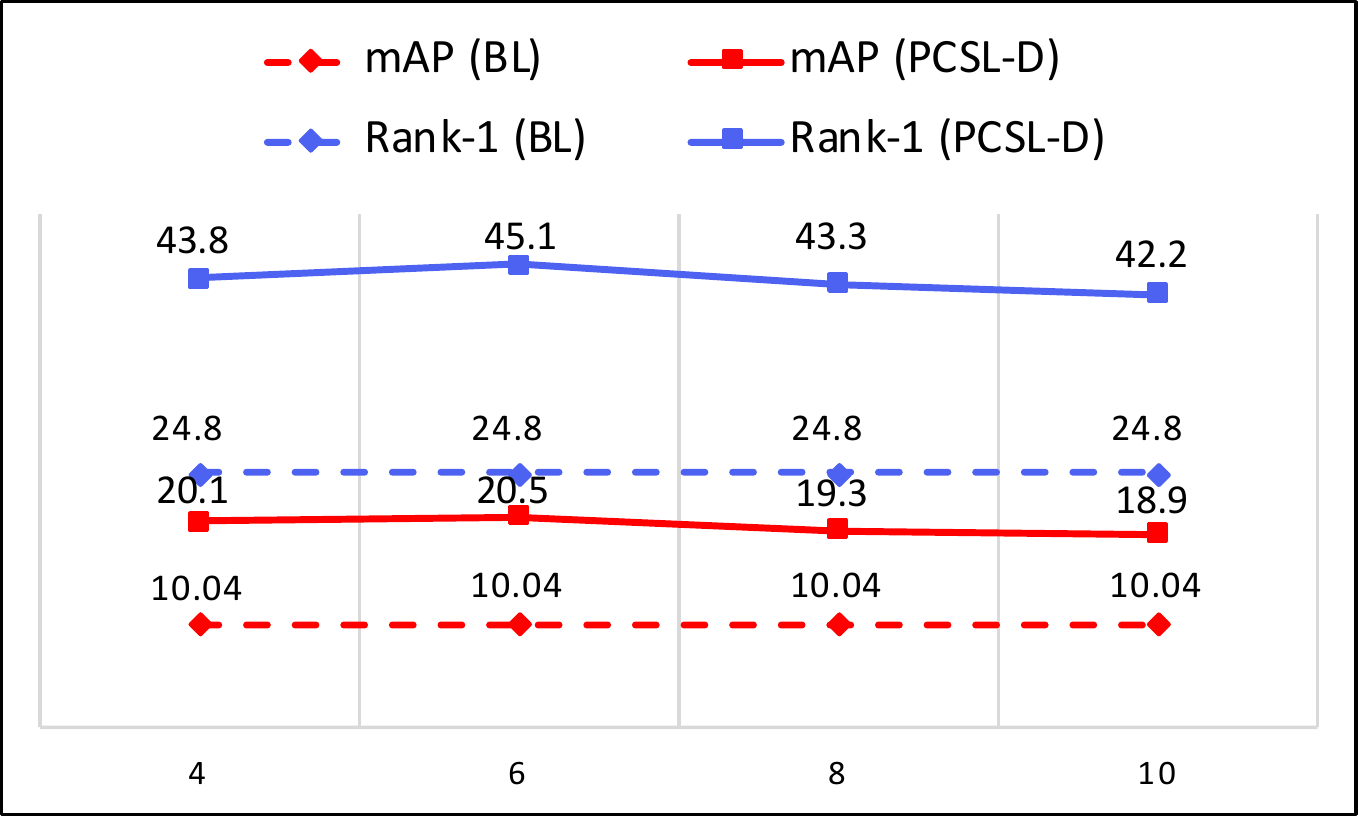}
}
\caption{Parameter analysis. In detail, a) shows curves for different $\lambda$ on Market1501; b) reports different $k$ on MSMT17}
\label{fig2}
   \vspace{-15pt}%
\end{figure}

\textbf{Parameter sensitivity.}
 To study the sensitivity of $\lambda$, which is used to trade off the intra-camera and inter-camera learning, we perform the experiments by PCSL-D on Market1501. As seen in Fig~\ref{fig2} (a), we obverse that mAP and the Rank-1 accuracy first increase and then decrease, and display as a bell-shaped curve. 
In addition, we conduct the experiments with different $k$ by PCSL-D, which is the parameter to select the number of the nearest persons, we sample the values in $\left\{4, 6, 8, 10\right\}$, and perform experiments by PCSL-D on MSMT17. The results are shown in Fig~\ref{fig2} (b), and we find that when $k$ is set as $6$, we can obtain the best results. 
 Compared with the baseline method (i.e., ``only $\mathcal{L}_{\mathrm{Intra}}$''), all settings can significantly improve the performance. Therefore, this further confirms the efficacy of our proposed framework. Finally, we set $\lambda$ and $k$ as $1$ and $6$ in all experiments for all datasets.

\textbf{Evaluation for the hard sample mining.}
In Eq.~(\ref{eq02}), we employ the hard-sample-mining scheme~\cite{hermans2017defense} to train our model, whose efficacy has been validated in the supervised person Re-ID task. To further confirm its efficacy in our setting, we compare it with the random-sample-mining scheme, i.e., generating triplets by randomly selecting positive and negative samples. We show the experimental results in Table~\ref{tab09}. As seen, this hard-sample-minging scheme can generate better triplets to optimize our model. In a training batch, if a triplet is easy, it could lead to ``$m+l(v_{a}^{i}) \leq 0$'' in Eq.~(\ref{eq01}), thus the loss of Eq.~(\ref{eq01}) equals $0$. This issue can be mitigated by mining the hard triplet.

\begin{table}[htbp]
  \centering
  \caption{Evaluation for the hard sample mining Market1501, DukeMTMC-reID (Duke) and MSMT17.}
    \begin{tabular}{|c|C{0.3cm}C{0.9cm}|C{0.3cm}C{0.9cm}|C{0.3cm}C{0.9cm}|}
    \toprule
    \midrule
    \multirow{2}[1]{*}{Method} & \multicolumn{2}{c|}{Market1501} & \multicolumn{2}{c|}{Duke} & \multicolumn{2}{c|}{MSMT17} \\
\cmidrule{2-7}          & mAP   & Rank-1 & mAP   & Rank-1 & mAP   & Rank-1 \\
    \midrule
    Random sample & 27.4  & 51.3  & 34.7  & 52.0  & ~6.1   & 16.0 \\
    Hard sample & \textbf{34.4} & \textbf{58.1} & \textbf{41.7} & \textbf{60.1} & \textbf{10.0} & \textbf{24.8} \\
    \bottomrule
    \end{tabular}
  \label{tab09}%
\end{table}%

\begin{figure}
\centering
\subfigure[PCSL-C]{
\includegraphics[width=4.1cm]{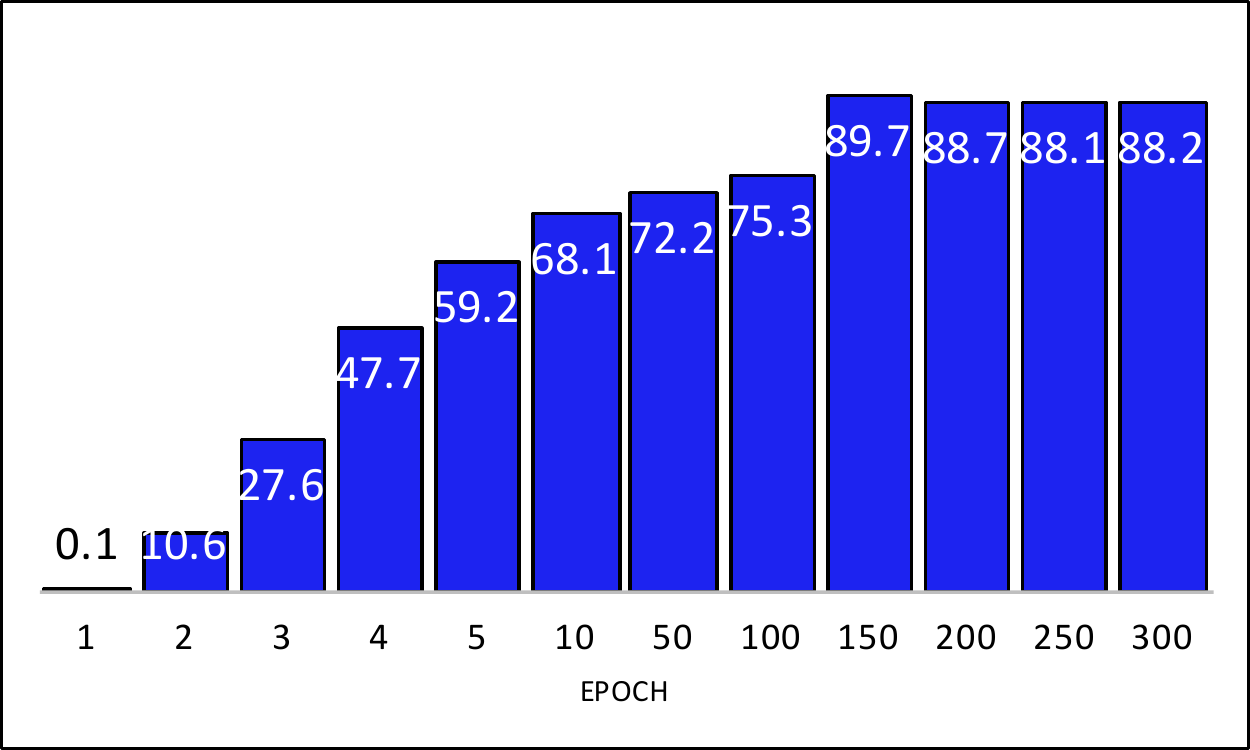}
}
\subfigure[PCSL-D]{
\includegraphics[width=4.1cm]{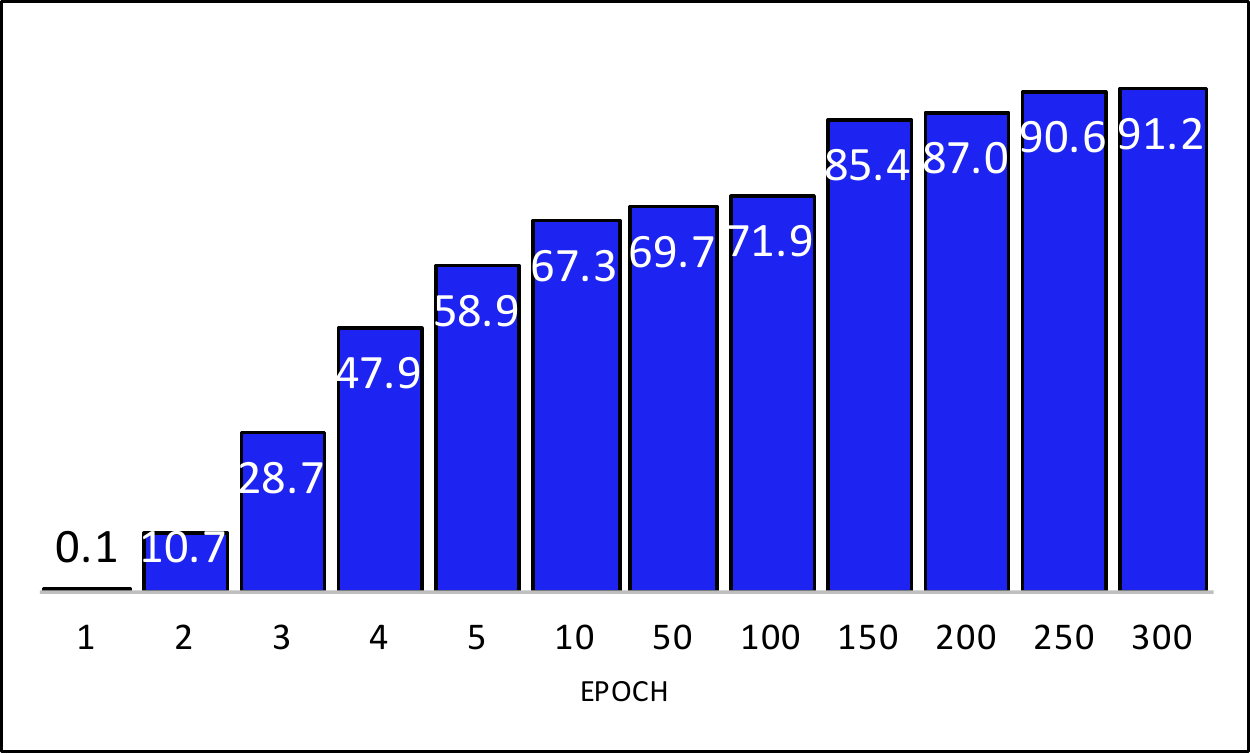}
}
\caption{Evaluation of affinity matrix by PCSL-C and PCSL-D on Market1501.}
\label{fig4}
\end{figure}

\begin{figure}
\centering
\includegraphics[width=6cm]{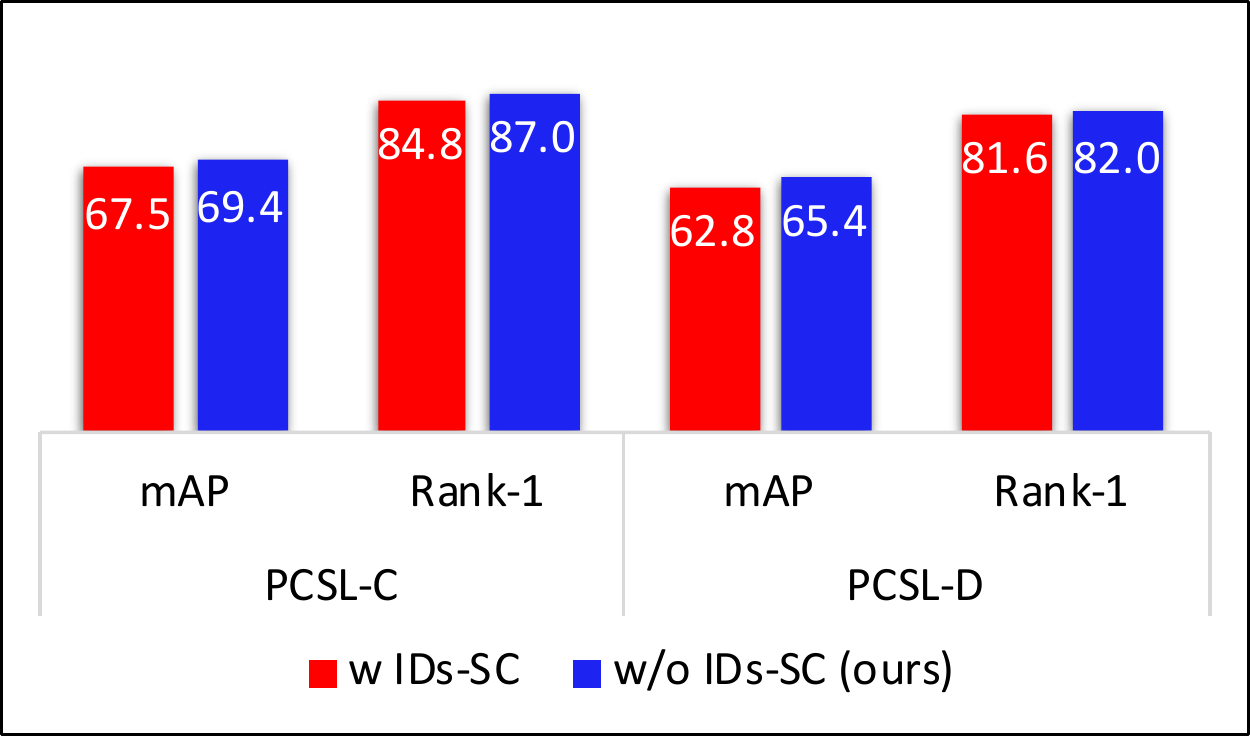}
\caption{Impact of the IDs in the same camera for the affinity matrix by PCSL-D and PCSL-C on Market1501, respectively. ``w/o IDs-SC'' and ``w IDs-SC'' indicate ``without IDs in the same camera'' and ``with IDs in the same camera'', respectively.}
\label{fig6}
\vspace*{-15pt}
\end{figure}

\textbf{Evaluation for affinity matrix.}
To observe the variation of the affinity matrix in the training process, we evaluate the quality of the affinity matrix by mAP. We hope that the true same identity persons are more similar (i.e., the corresponding element in the affinity matrix as shown in Fig.~\ref{fig9}) than the persons without the same IDs. Fig.~\ref{fig4} shows the results at different epochs on Market1501. As seen, with the epoch increasing, the affinity matrix has better performance, i.e., the feature representation has more discrimination. Particularly, when adding the unsupervised cross-camera learning at the $100$-th epoch, there is a large improvement from Epoch 100-150 in Fig.~\ref{fig4}. This confirms that the proposed framework can indeed produce progressive cross-camera soft-labels in the training course, as described in Section~\ref{SEC:UIL}.

\textbf{Impact of the intra-camera persons with respect to the affinity matrix.}
As discussed in Section~\ref{SEC:UIL}, since given intra-camera labels, we know the unique IDs in each cameras. Thus, we do not need to consider the personal relationships in the same camera due to the other within-camera persons could become some interference. This indeed brings some improvement in our task. The experimental result is shown in Fig.~\ref{fig6}. For example, mAP of ``w/o IDs-SC'' increases $1.9\%$ ($69.4$ vs. $67.5$) and $2.6\%$ ($65.4$ vs. $62.8$) over ``w IDs-SC'' for by PCSL-C and PCSL-D on Market1501, respectively.

\textbf{Evaluation of joint classification and discrimination tasks.}
In supervised person Re-ID, the effectiveness of simultaneously optimizing the classification and discrimination tasks is demonstrated in~\cite{zheng2018discriminatively}. We also use the same scheme in our framework, as reported in Table~\ref{tab07}. ``C'' is the classification task with the weighted cross-entropy loss, ``D'' denotes the discrimination task with the weighted triplet loss and  ``C+D'' indicates the joint classification and discrimination tasks. As seen in Table~\ref{tab07}, the joint scheme can slightly improve the performance of our task. However, the mAP on Duke is not improved. The main reason could be that soft-labels exist some noisy information that could remove the advantage of the joint classification and discrimination tasks. Compared with Market1501, Duke is a more difficult dataset for person Re-ID.

\begin{table}[htbp]
  \centering
  \caption{Performance of joint classification and discrimination tasks on Market1501 and DukeMTMC-reID (Duke). The best performance is \textbf{bold}.}
    \begin{tabular}{|c|cc|cc|}
    \toprule
    \midrule
    \multirow{2}[1]{*}{Method} & \multicolumn{2}{c|}{Market1501} & \multicolumn{2}{c|}{Duke} \\
\cmidrule{2-5}          & mAP   & Rank-1 & mAP   & Rank-1 \\
    \midrule
    C (ours) & 69.4 & 87.0 & 50.0  & 70.6 \\
    D (ours) & 65.4  & 82.0  & \textbf{53.5} & 71.7 \\
    \midrule
    C+D   & \textbf{69.9}  & \textbf{87.1}  & 50.3  & \textbf{72.6} \\
    \bottomrule
    \end{tabular}%
  \label{tab07}%
  \vspace*{-5pt}
\end{table}%

\begin{table}[htbp]
  \centering
  \caption{Comparison of different weighted schemes for the discrimination task on Market1501, DukeMTMC-reID (Duke) and MSMT17. The best performance is \textbf{bold}.}
    \begin{tabular}{|c|cc|cc|cc|}
    \toprule
     \midrule
    \multirow{2}[1]{*}{Method} & \multicolumn{2}{c|}{Market1501} & \multicolumn{2}{c|}{Duke} & \multicolumn{2}{c|}{MSMT17} \\
\cmidrule{2-7}          & mAP   & Rank-1 & mAP   & Rank-1 & mAP   & Rank-1 \\
    \midrule
    W & 63.6  & 81.6  & 51.0  & 68.9  & 19.1  & 43.0 \\
    AW (ours)  & \textbf{65.4} & \textbf{82.0} & \textbf{53.5} & \textbf{71.7} & \textbf{20.5} & \textbf{45.1} \\
    \bottomrule
    \end{tabular}%
  \label{tab08}%
  \vspace*{-5pt}
\end{table}%

\begin{figure}
\centering
\subfigure[Market1501]{
\includegraphics[width=4.1cm]{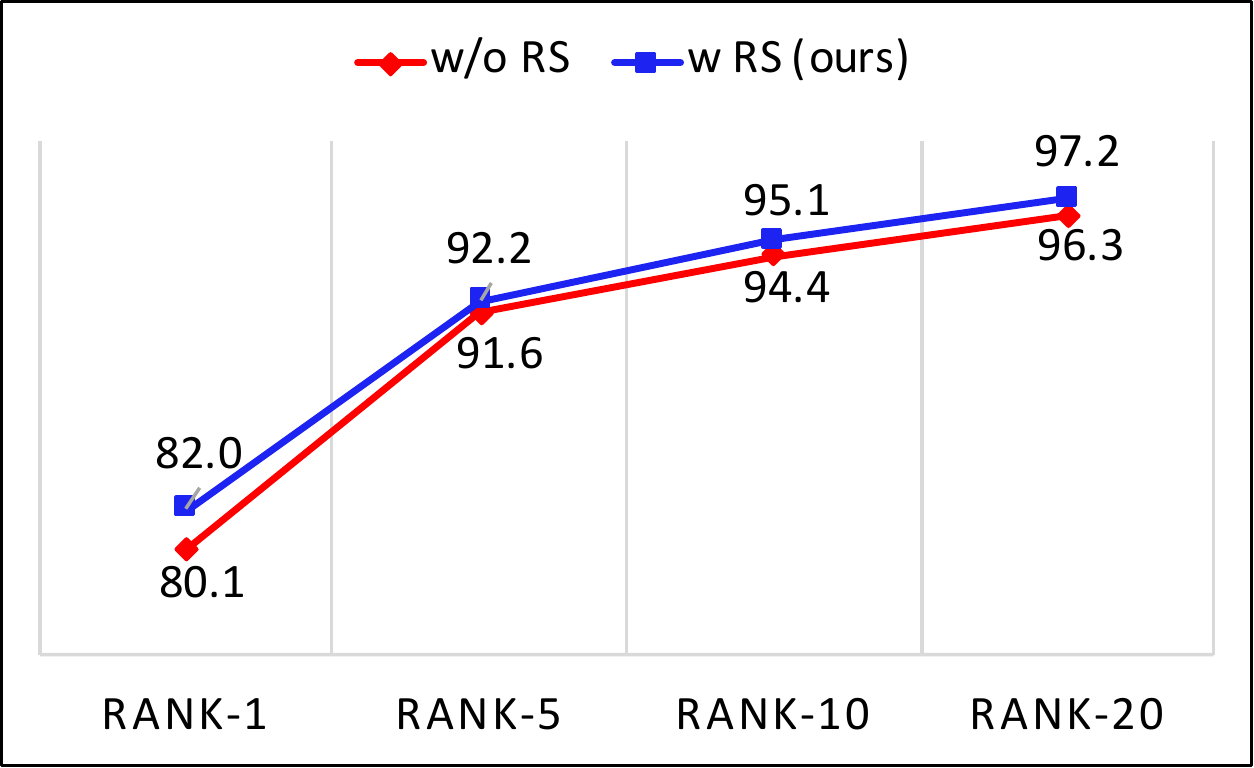}
}
\subfigure[MSMT17]{
\includegraphics[width=4.1cm]{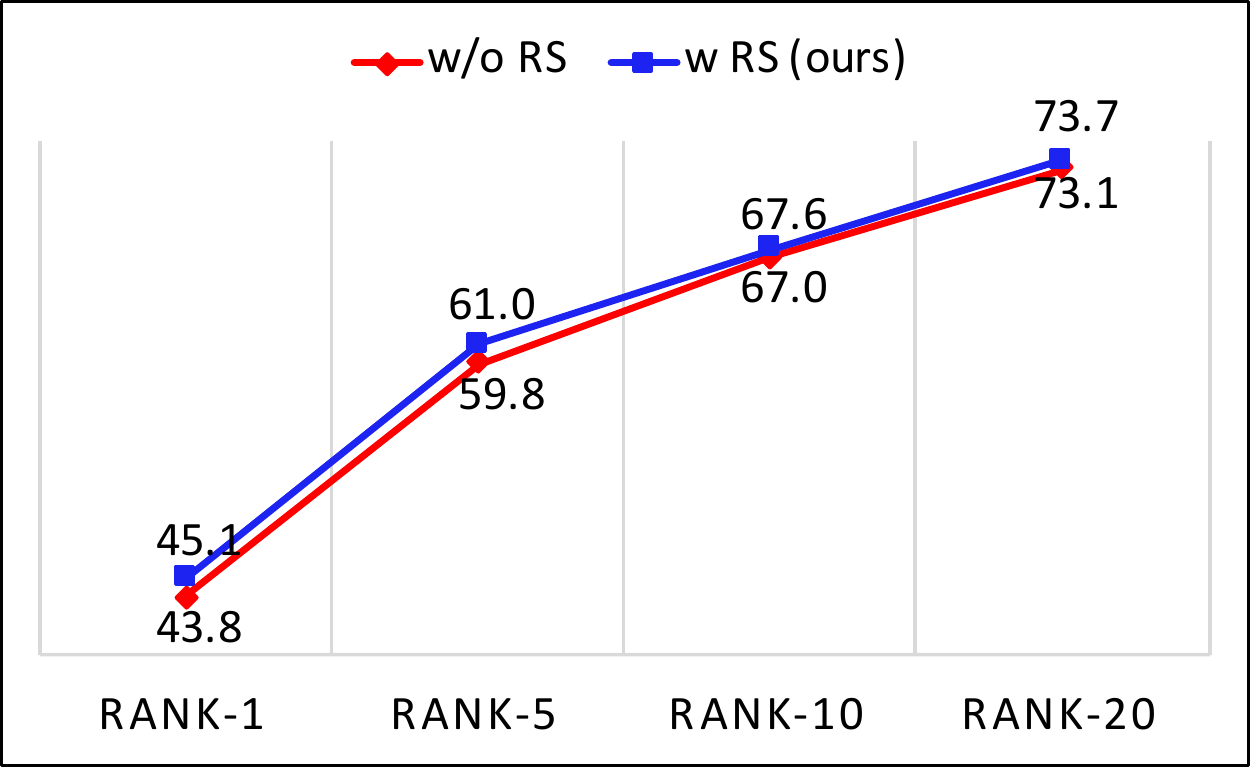}
}
\caption{Evaluation of the random sampling for PCSL-D on Market1501 and MSMT17.  ``w/o RS'' denotes ``without random sampling'' and ``w RS'' is ``with random sampling''.}
\label{fig7}
   \vspace{-15pt}%
\end{figure}

\textbf{Further evaluation for weighted triplet loss.}
In this part, we respectively evaluate the effectiveness of the average weighted scheme and the random sampling scheme in the weighted triplet loss, which are discussed in Section~\ref{SEC:UIL}. Firstly, for the average weighted scheme (AW), we directly utilize the weights based on the cross-camera person similarities (W) as the comparison. The results are shown in Table~\ref{tab08}. As seen, ``AW'' consistently outperforms ``W'' on all datasets, which demonstrates the effectiveness of the average weight scheme, as analyzed in Section~\ref{SEC:UIL}. Secondly, for the positive sample selection, we also compare the random sampling scheme with the method of selecting the nearest persons with the anchor. As seen in Fig.~\ref{fig7}, the random sampling scheme could select some hard positive samples, while selecting the most similar persons means to choose easier positive samples. Thus, it is consistent with the analysis in the literature~\cite{hermans2017defense}, i.e., hard positive samples can help deep models to enhance the generalization ability in person Re-ID.
As analyzed in Section~\ref{SEC:UIL}, we also compare our loss in Eq.~(\ref{eq09}) with the Adaptive Weighted Triplet Loss (AWTL)~\cite{Ristani_2018_CVPR}, which aims to give a large penalty to the hardest positive and negative samples by assigning a large weight. It is developed to handle the supervised person Re-ID task and needs the true label information for all training samples. As seen in Table~\ref{tab10}, AWTL has poorer performance than our method. In our setting, there are only true labels for the within-camera images while the cross-camera labels are generated by our proposed method, which could not be completely correct in the estimating of these soft-labels (i.e.,  the chosen positive sample may be the true negative sample), especially for these hardest positive samples. In this case, if we simply give a large penalty to these hard samples, it will result in poor performance.

\begin{table}[htbp]
  \centering
  \caption{Evaluation of Adaptive Weighted Triplet Loss (AWTL) with our generated soft-labels on Market1501, DukeMTMC-reID (Duke) and MSMT17. The best performance is \textbf{bold}.}
    \begin{tabular}{|c|cc|cc|cc|}
    \toprule
    \midrule
    \multirow{2}[1]{*}{Method} & \multicolumn{2}{c|}{Market1501} & \multicolumn{2}{c|}{Duke} & \multicolumn{2}{c|}{MSMT17} \\
\cmidrule{2-7}          & mAP   & Rank-1 & mAP   & Rank-1 & mAP   & Rank-1 \\
    \midrule
    AWTL~\cite{Ristani_2018_CVPR}  & 61.7  & 79.1  & 51.8  & 71.2  & 18.8  & 41.4 \\
    Ours  & \textbf{65.4} & \textbf{82.0} & \textbf{53.5} & \textbf{71.7} & \textbf{20.5} & \textbf{45.1} \\
    \bottomrule
    \end{tabular}%
  \label{tab10}%
\end{table}%

\textbf{Analysis of the robustness of our method.}
To validate the robustness of our method, we conduct two different experiments: i) Removing overlapping images in the original training set to keep images of only one camera for 25\%, 50\% and 75\% identities. ii) Adding the images of persons that only appear under one camera into the original training set. Also, we explain why these two different experiments are conducted.

In the first experiment, by removing the overlapping images in the training set, we find that it will generate an unfair comparison with baseline, which is obtained with all of the 12936 training images on Market1501. This is because fewer training images can be retained when keeping images of only one camera for 25\%, 50\% and 75\% identities, as shown in Table~\ref{tab12} in this response letter. Note that in this table ``single-25/50/75'' denotes 25/50/75\% persons appear under one camera. We can see the respective number of training samples decreases from 12936 to 10436, 7895 and 5367.

To avoid an unfair comparison, we have to remove the effect of unequal number of training images. Therefore, we reconstruct the original training set to ensure it to have almost the equal number of training images in each case as follows: \\
i) For the setting of ``single-25/50/75'', we randomly select 25\%, 50\% or 75\% persons from the 751 persons. For each selected person, we randomly choose one camera to retain all images of this person from it and remove all images of this person from other cameras. 
ii) For each corresponding ``multiple-all'', we randomly select one person from the 751 persons. If the chosen person appears under three or more cameras (i.e., \#camera $\geq$ 3), we will randomly select one camera to remove all images of this person from it. We iteratively conduct the above two steps until the number of retained images is almost equal to the number of training images obtained in the above ``single-25/50/75'' cases. In doing so, we can largely remove the effect of unequal number of training images when conducting the comparison.

To accumulate statistics, we repeat the above construction of these training sets by $5$ times and report the results averaged on them in Table~\ref{tab12}. As seen, the proposed method works better in the setting of``single-25/50/75'' than ``multiple-all'' in most cases, especially with the weighted cross-entropy loss based on the classification task (i.e., PCSL-C)~\footnote{At the same time, it can be well expected that the proposed method in these settings cannot compete with the original setting in which the number of training images is much larger.}. This result preliminarily demonstrates the robustness of the proposed method when a sufficient number of persons only have images under a single camera. We further look into this result to gain more understanding on why the proposed method works better, and this leads to the observation presented in Table~\ref{tab15}. As shown, the setting of ``single-25/50/75'' contains more persons whose images appear under 5 or 6 cameras than the setting of ``multiple-all''. This can be expected because the former is designed to make 25\%, 50\% or 75\% persons have images under one camera only, which provides room for the remaining persons to appear under more cameras (please be reminded of the constraint that the two settings need to have almost equal number of training images). The presence of more images appearing under 5 or 6 cameras offers stronger cues for the proposed method to figure out the relationships across different cameras. This is why it can work better.


\renewcommand{\cmidrulesep}{0mm}  
\setlength{\aboverulesep}{0mm} 
\setlength{\belowrulesep}{0mm} 
\setlength{\abovetopsep}{0cm}  
\setlength{\belowbottomsep}{0cm}
\begin{table*}
  \centering
  \caption{Comparison between baseline and the proposed method in the cases of keeping images of only one camera for 25\%, 50\% and 75\% identities. Note ``single-25'' denotes 25\% persons appear under one camera only. ``multiple-all'' means that all persons appear under multiple cameras (i.e., \#camera $\geq$ 2). ``original'' indicates the baseline using the original training setting on Market1501.}
    \begin{tabular}{|c|c|c|cc|cc|}
    \toprule
    \midrule
    \multirow{2}[1]{*}{Setting} & \multirow{2}[1]{*}{\#training image} & \multirow{2}[1]{*}{\#training ID} & \multicolumn{2}{c|}{PCSL-C} & \multicolumn{2}{c|}{PCSL-D} \\
\cmidrule{4-7}          &   &    & mAP   & Rank-1 & mAP   & Rank-1 \\
    \midrule
    singel-25 & 10436 & 751 & 59.3  & 76.1  & 56.3  & 73.3 \\
    mulitple-all & 10395 & 751 & 58.3  & 75.3  & 55.3  & 71.9 \\
    \midrule
    singel-50 & 7895& 751 & 54.3  & 72.9  & 48.2  & 66.9 \\
    mulitple-all & 7894 & 751 & 50.5  & 69.9  & 49.9  & 68.7 \\
    \midrule
    singel-75 & 5367 & 751 & 44.4  & 65.6  & 42.6  & 64.1 \\
    mulitple-all & 5368 & 751 & 41.1  & 62.1  & 42.7  & 63.4 \\
    \bottomrule
    \midrule
    original & 12936 & 751 & 66.9 & 80.1 & 64.2 & 78.2 \\
    \bottomrule
    \end{tabular}%
  \label{tab12}%
\end{table*}%

\begin{table*}
  \centering
  \caption{The statistics of the constructed training set in each case. \#ID-2CAM denotes the number of persons who appear under 2 cameras. Note that the mean results are given in this table.}
    \begin{tabular}{|c|cccccc|}
    \toprule
    \midrule
    Setting & \#ID-1CAM  & \#ID-2CAM    & \#ID-3CAM     & \#ID-4CAM      & \#ID-5CAM     & \#ID-6CAM \\
    \midrule
    single-25 & 188.0 & 36.4  & 81.2  & 156.2 & 226.6 & 62.6 \\
    multiple-all & 0.0   & 189.0 & 196.6 & 212.4 & 126.6 & 26.4 \\
    \midrule
    single-50 & 376.0 & 25.0  & 50.8  & 106.6 & 153.0 & 39.6 \\
    multiple-all & 0.0   & 396.0 & 238.6 & 95.0  & 19.4  & 2.0 \\
    \midrule
    single-75 & 563.0 & 14.2  & 25.2  & 52.4  & 76.6  & 19.6 \\
    multiple-all & 0.0   & 751.0 & 0.0   & 0.0   & 0.0   & 0.0 \\
    \bottomrule
    \end{tabular}%
  \label{tab15}%
   \vspace*{-15pt}%
\end{table*}%

Although the above result is positive for the proposed method, we are not fully satisfied with it because, after all, the two settings in comparison have different numbers of persons under each camera. It will be good to conduct additional experiments in which such a discrepancy can be considerably removed.
This leads us to the second experiment as follows. In this experiment, we reconstruct the training set on Market1501 to further validate the robustness of our method, as shown in Table~\ref{tab13}. Different from the first experiment, we now take an auxiliary dataset from the original testing dataset of Market1501. In other words, we split the original testing set of Market1501 into the auxiliary set and a new testing set. This auxiliary set is used to generate the images of persons that appear under only one camera. Note that we do not take the auxiliary set from any other person Re-ID datasets because this could bring in unnecessary domain shift between two datasets.

\begin{table}[htbp]
  \centering
  \caption{Dataset partition on Market1501. Note that ``\#training ID'', ``\#auxiliary ID'' and ``\#testing ID'' denote the number of persons in the training, auxiliary and testing sets, respectively.}
    \begin{tabular}{|c|c|c|c|}
    \toprule
    \midrule
    Dataset & \#training ID & \#auxiliary ID & \#testing ID \\
    \midrule
    Original (used in baseline) & 751   & 0     & 750 \\
    \midrule
    Newly constructed  & 751   & 300   & 450 \\
    \bottomrule
    \end{tabular}%
  \label{tab13}%
\end{table}%

In our experiment, we add 100, 200 and 300 IDs into the original training set, respectively. For each ID in the auxiliary set, images under a single camera are randomly selected, which means these IDs will have no cross-camera image pairs in the training set. Thus, the setting is good for validating the robustness of our method because the number of persons under a given number of cameras maintains to be the same, except that for the single camera. In this case, the images that only appear under one camera can be seen as ``noisy data'' because they cannot help our method to explore the relationships across different cross-camera persons. We report the experimental results in Table~\ref{tab14}. As seen, i) our method is robust against the number of noisy data when using the weighted cross-entropy loss based on the classification task (i.e., PCSL-C); ii) Using the weighted cross-entropy loss is more robust than the weighted triplet loss (i.e., PCSL-D). The main reason is that the weighted triplet loss is obtained by directly computing the distance loss, which is more sensitive than the classification loss for the ``noisy data'' due to selecting some false positive samples to train the model. 

\begin{table}[htbp]
  \centering
  \caption{Evaluation of the robustness of our method on the newly constructed training and testing sets on Market1501. ``\#training image'' denotes the number of training images. For ``\#multiple+\#single'', ``\#multiple'' denotes the number of persons appearing under multiple cameras (\#camera $\geq$ 2) and ``\#single'' denotes the number of persons only appearing under a single camera.}
    \begin{tabular}{|c|c|C{0.3cm}C{0.9cm}|C{0.3cm}C{0.9cm}|}
    \toprule
    \midrule
    \multirow{2}[1]{*}{\#training image} & \multirow{2}[1]{*}{\#multiple+\#single} & \multicolumn{2}{c|}{PCSL-C} & \multicolumn{2}{c|}{PCSL-D} \\
\cmidrule{3-6}          &       & mAP   & Rank-1 & mAP   & Rank-1 \\
    \midrule
    12936 & 751+0 & 66.86 & 80.08 & 64.21 & 78.19 \\
    \midrule
    13393 & 751+100 & 67.14 & 80.28 & 62.49 & 76.25 \\
    13835 & 751+200 & 67.42 & 79.43 & 61.74 & 76.20 \\
    14364 & 751+300 & 67.41 & 80.28 & 61.40 & 74.07 \\
    \bottomrule
    \end{tabular}%
  \label{tab14}%
     \vspace{-15pt}%
\end{table}%
\section{Conclusion}\label{s-conclusion}
In this paper, we focus on the semi-supervised person Re-ID case, which only provides the within-camera labels but not cross-camera labels. To well address this case, we put forward a progressive cross-camera soft-label learning framework, which can obtain progressive cross-camera soft-labels and incrementally improved feature representations by the alternative learning way. 
Extensive experiments show that the proposed method has a completive performance when compared with deep supervised methods. Furthermore, the efficacy of all components in the proposed framework is validated by ablation study. As reported in Fig.~\ref{fig4}, the quality of the affinity matrix still can be further promoted. In future work, we will explore the new method to obtain a better affinity matrix and more stable cross-camera soft-labels for the semi-supervised person Re-ID. 


%
%

\ifCLASSOPTIONcaptionsoff
  \newpage
\fi

\bibliographystyle{IEEEtran}
\bibliography{sigproc}
\end{document}